\documentclass[pdflatex,sn-nature]{sn-jnl}

\usepackage{graphicx}
\usepackage{amsmath,amssymb,amsfonts}
\usepackage{booktabs}
\usepackage{multirow}
\usepackage{xcolor}
\usepackage{textcomp}
\usepackage{manyfoot}
\begin{document}

\title[When High Capacity Becomes Undeployable]{When High Capacity Becomes Undeployable: Pore-Filling Constraints for Multi-Objective Optimization of Direct Air Capture Sorbents}

\author*[1]{\fnm{Yujie} \sur{Liu}}\email{liuyj@sustech.edu.cn}
\author[1]{\fnm{Xuehua} \sur{Shen}}
\author*[1]{\fnm{Zuotai} \sur{Zhang}}\email{zhangzt@sustech.edu.cn}

\affil*[1]{\orgname{Southern University of Science and Technology}, \orgaddress{\city{Shenzhen}, \postcode{518055}, \state{Guangdong}, \country{China}}}

\abstract{Surrogate-based optimization of amine-impregnated sorbents can favour high-capacity formulations that violate pore-volume constraints, opening a gap between nominal and physically feasible performance. Yet multi-objective Bayesian optimization (MOBO) for direct air capture (DAC) treats formulation search as unconstrained. We encode the pore-filling boundary as an explicit outcome constraint, $c(x)=\tanh(1-\hat{\phi})-\tau$, and use it as a diagnostic instrument on a DOI-traceable benchmark of 618 solid-amine formulations from 43 publications, expanding a public dataset by 27 records. Unconstrained MOBO attains higher nominal capacity ($3.639 \pm 0.815$ vs.\ $2.964 \pm 0.728$ mmol\,g$^{-1}$, $p=1.0\times10^{-11}$), but across 100 seeds $63.2\%$ of its Pareto-optimal candidates violate the boundary versus $39.3\%$ under the constraint: unconstrained search is rewarded outside the pore-confined regime. The constraint steers the front to lower loadings (55.8 vs.\ 65.2 wt\%) and larger pores. The boundary is corroborated once apparatus effects are harmonized hierarchically: 77.8\% of kinetic variance is between-publication, and pore filling carries a significant kinetic penalty ($p=0.002$). Standard hypervolume over-credits feasibility-agnostic methods by 13.9--16.1\% (vs.\ 11.1\% constrained), and three surrogate-based analyses---the constrained front, a sensible-heat regeneration-energy proxy, and harmonized kinetics---point to a conservative candidate design region at pore filling $\hat{\phi}\approx0.4$--$0.5$ (45--55 wt\% on high-pore-volume silicas), pending experimental validation. The benchmark and analysis outputs accompany this submission.}

\keywords{direct air capture, solid amine sorbents, multi-objective Bayesian optimization, pore filling, feasibility constraint, literature mining}

\maketitle

\section{Introduction}\label{sec:intro}

Direct air capture (DAC) of CO$_2$ is increasingly recognised as a necessary component of net-zero portfolios, and solid sorbents bearing amine functionalities are among the most mature material classes for the dilute conditions (ca.\ 400 ppm CO$_2$) that define the problem \cite{sanzperez2016,jones2011,bui2018,keith2018}. Within this class, amine-impregnated (``class 1'') sorbents---polyamines physically loaded into porous supports---offer high capacities and simple synthesis \cite{goeppert2011,choi2011,gebald2011}, and their formulation space (support, amine identity and blend, loading, pore architecture) is large, combinatorial, and only sparsely explored \cite{shi2020,didas2015}.

A persistent theme in this literature is the tension between capacity and accessibility. As amine loading rises, capacity per gram of sorbent increases until the amine phase begins to obstruct the pore network; beyond that point diffusion slows, amine efficiency falls, and additional loading buys little or nothing \cite{bollini2011,sayari2010,serna2010,li2010}. The physical origin is simple pore-volume conservation: the volume of impregnated amine cannot exceed the pore volume of the dry support without occupying external surface or blocking pore mouths. This boundary is well understood qualitatively, yet it is almost never made explicit in computational formulation search.

Multi-objective Bayesian optimization (MOBO) has become a standard tool for navigating materials formulation spaces \cite{shahriari2016,frazier2018,daulton2020}, and surrogate models trained on literature-mined datasets are increasingly used to accelerate sorbent design. Constrained Bayesian optimization is itself well established---expected improvement under unknown or classifier-modelled constraints, and scalable constrained variants \cite{gelbart2014,gramacy2011,gardner2014,letham2019,eriksson2021}---but the constraints in that literature are generic black-box safety or validity functions; encoding a pore-level physical conservation law as the constraint in sorbent formulation search, and using it to quantify how much of unconstrained optimization's apparent advantage lies outside the pore-confined design regime, has not, to our knowledge, been explicitly formulated in DAC sorbent formulation search. Applied naively, however, such optimization rewards exactly the failure mode described above: the nominal capacity maximum of any surrogate trained on published data sits near or beyond the pore-filling limit, because that is where the highest reported capacities lie. The optimizer is then optimizing formulations for which the pore-confined description on which the surrogate is built no longer applies. Standard evaluation metrics compound the problem: hypervolume and best-found capacity credit every proposed candidate, including ones geometrically inconsistent with pore-confined impregnation.

Here we show that unconstrained optimization of DAC sorbents is systematically biased toward pore-filling-inconsistent formulations. We correct this bias by encoding a pore-level feasibility boundary into batch MOBO, introduce feasible hypervolume as a deployability-aware evaluation metric, and identify a conservative design window that remains preferred across raw optimization, harmonized kinetics, and an energy proxy. In detail, our contributions, in order of priority, are: (i) a physics-derived feasibility constraint $c(x)=\tanh(1-\hat{\phi})-\tau$ with amine-specific densities, integrated into qLogEHVI batch optimization \cite{balandat2020,ament2023}; (ii) a demonstration, across 100 seeds, that the nominal capacity advantage of unconstrained MOBO is concentrated in geometrically inconsistent candidates, while the constraint steers the front to lower loadings and larger pores; (iii) evidence that feasibility-blind metrics systematically over-reward feasibility-agnostic methods (13.9--16.1\% hypervolume inflation, versus 11.1\% under the constraint), with a deployability-aware alternative (feasible hypervolume); and (iv) a deployable formulation window on which three surrogate-based analyses---the constrained front, a sensible-heat regeneration-energy proxy analysis, and kinetics harmonized for apparatus effects---converge. Throughout, ``deployability'' refers specifically to geometric consistency with the pore-volume boundary---a necessary condition for pore-confined impregnation, not a complete engineering assessment of intraparticle amine distribution, mass transfer, cycling stability, or process economics. Supporting these analyses, we assemble a DOI-traceable benchmark of 618 solid-amine formulations with paired capacity and kinetics from 43 publications, built on a public dataset \cite{wang2025scidata} and extended by independent curation, including a quantified audit of which literature reporting practices prevent data pooling, and we corroborate the boundary itself: its kinetic penalty emerges after hierarchical harmonization (77.8\% of kinetic variance is between-laboratory; $\hat{\phi}$ effect $p=0.002$), amine efficiency dips significantly in the transition zone, and its sign is reproducible from formulation descriptors when pore volume is unreported (an internal consistency check, LOPO AUC $=0.93$).

\section{Results}\label{sec:results}

\subsection{A DOI-traceable benchmark and the limits of literature interoperability}\label{sec:benchmark}

The benchmark builds on the public dataset of Wang \emph{et al.} \cite{wang2025scidata}, which compiles 1{,}336 manually collected data rows from 52 peer-reviewed publications on amine-impregnated solid adsorbents---the most extensively reviewed sorbent class for DAC \cite{sanzperez2016,shi2020,berger2011}. After normalizing DOI strings, we retained entries measured under dry conditions (relative humidity 0\%) with both an equilibrium capacity and a kinetics metric ($t_{90}$, or $t_{50}$ convertible to $t_{90}$ assuming first-order kinetics, $t_{90}=t_{50}\cdot\ln 10/\ln 2$), and complete formulation descriptors; exact duplicates were removed. An independent stage-0 curation effort contributed 68 additional candidate entries, of which only 27 could be merged: grafted (``class 2'') sorbents report no impregnation loading, so $\hat{\phi}$ is undefined; breakthrough protocols report $t_5$/$t_{50}$ times not convertible to batch $t_{90}$; and linear-driving-force constants are not transferable across apparatus classes. Reporting practice, not data quality, excluded the majority---a finding we return to in the Discussion. The final benchmark contains \textbf{618 entries from 43 publications}: a 591-entry core library filtered from that dataset (507 reported $t_{90}$ values plus 84 converted from $t_{50}$) and 27 merged stage-0 entries carrying per-row provenance flags. Operating conditions are heterogeneous by design: 59.7\% of core entries are flue-gas-scale measurements at 100 vol\% CO$_2$, temperatures span $-20$ to $120\,^\circ$C, and 146 entries fall in a DAC-relevant window (20--35\,$^\circ$C, $\leq 600$ ppm). Cross-validated capacity error inside this DAC window is comparable to the high-concentration bands under random splits; under the stricter publication-grouped split its capacity error remains the lowest of the three bands, although the explained variance decays (Supplementary Section S4). The optimization conclusions below are therefore statements about optimizer behaviour on the literature-calibrated surrogate landscape, with external validity bounded by the grouped cross-validation. Conditions enter the models as features (Methods), and all reported optimizations are pinned to the DAC deployment point (25\,$^\circ$C, 0.04 vol\% CO$_2$, dry, 100 mL\,min$^{-1}$). Every entry carries a DOI linking to the original publication; 84.5\% of entries satisfy the pore-filling boundary ($\hat{\phi}\leq1$, median $\hat{\phi}=0.632$).

\subsection{Empirical status of the pore-filling constraint}\label{sec:validation}

Three analyses probe the feasibility boundary itself. \textbf{(a) Learnability---an internal consistency check, not independent validation.} Because $\hat{\phi}$ is a closed-form function of loading, amine density, and pore volume, and because pore volume enters our pipeline as a per-support lookup, a model given support identity, amine, loading, BET area, and pore diameter could in principle re-derive the boundary arithmetically. We therefore report learnability only as a consistency check with one practical consequence---formulations can be screened against the boundary without a measured pore volume---not as physical validation. Under leave-one-publication-out cross-validation the feasibility \emph{sign} is reproduced with AUC $=0.93$ (linear, GP-RBF, and gradient-boosted models agree within 0.01; continuous $R^2=0.57$--$0.60$). That this skill is the support-to-pore-volume mapping rather than new physics is shown by the audit of Supplementary Section S3: removing support identity (leave-one-support-out) destroys predictive skill (median $R^2\approx-0.7$). The non-trivial empirical corroborations are (b) and (c). \textbf{(b) Kinetic corroboration via hierarchical harmonization.} A naive pooled analysis is uninformative: within temperature$\times$CO$_2$-concentration strata, $\hat{\phi}$ does not discriminate kinetically penalized entries (AUC $=0.505$). A hierarchical mixed-effects model of $\log_{10}(1/t_{90})$ on temperature, log CO$_2$ concentration, loading, and $\hat{\phi}$, with a random intercept per publication ($n=591$, 40 groups; the 591-entry core spans 40 of the 43 benchmark publications---the 27 merged stage-0 entries contribute the remaining three), attributes \textbf{77.8\% of residual variance to publication-level variance} (between-publication differences in apparatus, protocols, and material families; ICC $=0.778$; marginal $R^2=0.35$ vs.\ conditional $R^2=0.86$). Controlling for this heterogeneity, $\hat{\phi}$ has a significant negative fixed effect on log-rate ($\beta=-0.184$, 95\% CI $[-0.303, -0.065]$, $p=0.002$): higher pore filling slows uptake. A within-publication analysis corroborates this: across 19 publications with $\geq8$ entries spanning a $\hat{\phi}$ range, the median Spearman correlation between $\hat{\phi}$ and log-rate is $-0.279$ (16/19 negative; 7/19 individually significant). Two robustness checks support this association: a random-slope specification retaining a per-publication $\hat{\phi}$ slope leaves the fixed effect significant ($\beta=-0.420$, 95\% CI $[-0.737,-0.102]$, $p=0.010$), and despite the structural $\hat{\phi}$--loading association (Spearman $\rho=0.73$) the fixed-design variance inflation factors are $\leq1.5$, so the effect is not an alias of loading (Supplementary Section S4). Because the kinetic surrogate itself does not generalize across publications (Supplementary Section S4), we read these results strictly as association-level evidence and use kinetics only as a harmonized relative indicator. The pore-filling--kinetics trade-off is thus real but masked in pooled literature data by apparatus heterogeneity. \textbf{(c) Amine efficiency.} Apparent amine efficiency (capacity per unit loading) dips significantly in the transition zone $0.9<\hat{\phi}\leq1.1$ (median 0.031 vs.\ 0.039 mmol\,g$^{-1}$ per wt\% for $0.7<\hat{\phi}\leq0.9$; Mann--Whitney $p=0.005$), consistent with incipient pore blocking; for $\hat{\phi}>1.1$ efficiency recovers relative to the transition zone (median 0.048 vs.\ 0.031; $p=0.001$), indicating breakdown of the geometric assumptions and bounding the constraint's domain of validity. The recovery should not be read as evidence that individual overfilled materials escape pore blocking: it is equally consistent with amine accumulating outside the pore network as an external phase, with between-study differences in how loading and pore volume are measured, and with the composition of the small high-filling subsample. We therefore treat the upturn strictly as marking where the geometric model stops being informative.

\subsection{Constrained optimization trades nominal capacity for deployability}\label{sec:main}

\begin{figure}[t]
\centering
\includegraphics[width=0.85\textwidth]{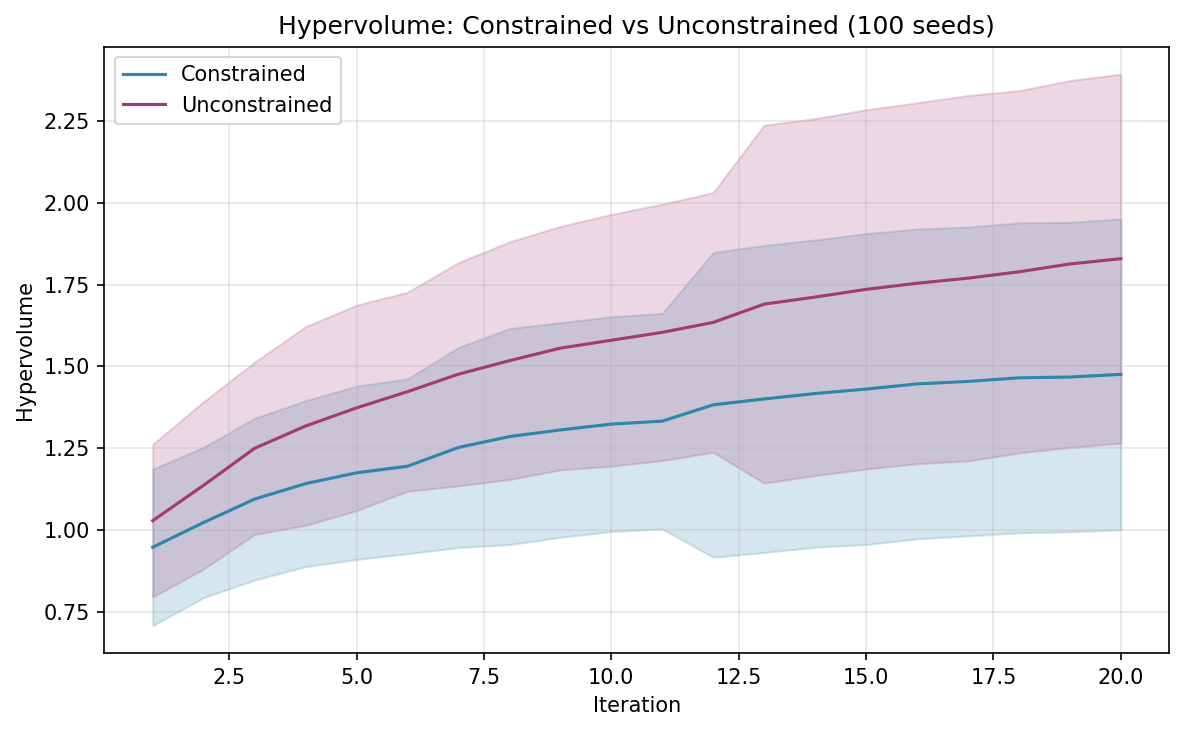}
\caption{\textbf{Unconstrained MOBO's hypervolume lead is established early and persists, but much of it is carried by geometrically inconsistent candidates} (Section~\ref{sec:feashv}). Hypervolume trajectories over 100 seeds (20 iterations, batch $\leq4$; shared reference point computed from pooled objective vectors).}\label{fig:hv}
\end{figure}

Table~\ref{tab:main} and Fig.~\ref{fig:hv} summarize final (best-found) capacity distributions over 100 independent seeds on the merged benchmark. The unconstrained optimizer attains a higher mean final capacity than the constrained optimizer ($3.639\pm0.815$ vs.\ $2.964\pm0.728$ mmol\,g$^{-1}$; Wilcoxon signed-rank $p=1.0\times10^{-11}$). Section~\ref{sec:front} shows that much of this advantage is carried by candidates that violate the pore-filling boundary. Random search is statistically indistinguishable from the constrained optimizer in nominal capacity ($2.944\pm0.533$, $p=0.78$), while single-objective expected improvement attains the highest nominal capacity ($3.826\pm0.823$, $p=1.3\times10^{-12}$) with neither kinetics nor deployability guarantees---underscoring that the constraint's value lies in deployability, not in capacity-seeking efficiency. Because nominal capacity credits geometrically inconsistent candidates, we treat a deployability-aware metric (feasible hypervolume, Section~\ref{sec:feashv}) as a co-primary endpoint throughout (per-seed details and further sensitivity analyses: Supplementary Section S4).

\begin{table}[t]
\centering
\caption{Final capacity distributions over 100 independent seeds (20 iterations, batch $\leq4$) on the merged 618-entry benchmark. Wilcoxon signed-rank tests are against the constrained MOBO.}\label{tab:main}
\begin{tabular}{lccc}
\toprule
Method & Final capacity (mmol\,g$^{-1}$) & Std dev & $p$-value \\
\midrule
Constrained MOBO (ours) & 2.964 & 0.728 & -- \\
Unconstrained MOBO & 3.639 & 0.815 & $1.0\times10^{-11}$ (sig.) \\
Single-objective EI & 3.826 & 0.823 & $1.3\times10^{-12}$ (sig.) \\
Random search & 2.944 & 0.533 & 0.78 (n.s.) \\
\bottomrule
\end{tabular}
\end{table}

\subsection{Structural shift of the Pareto front}\label{sec:front}

\begin{figure}[t]
\centering
\includegraphics[width=0.9\textwidth]{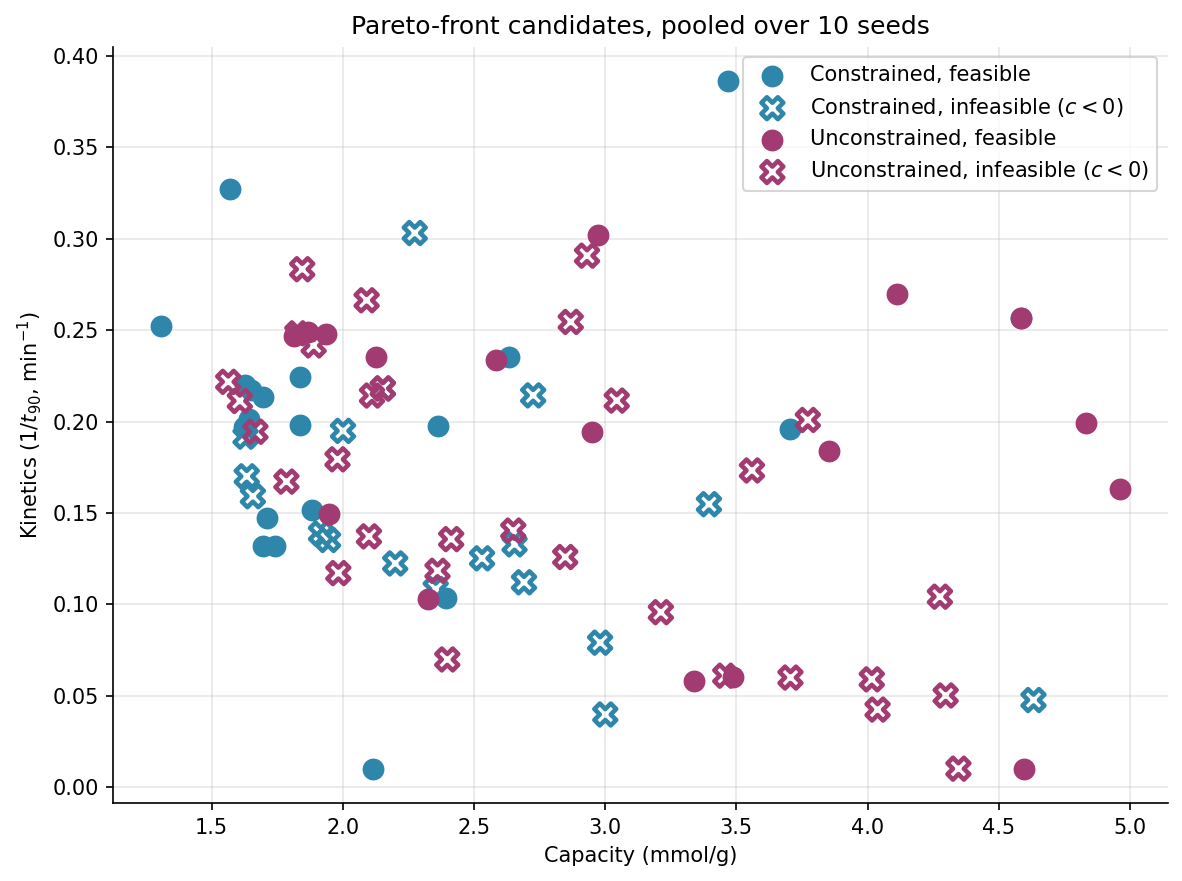}
\caption{\textbf{The constraint reduces systematic drift toward overfilled formulations.} Pareto-front candidates from an illustrative 10-seed pool (filled circles: feasible, $c(x)\geq0$; open crosses: predicted infeasible, $c(x)<0$). Over the full 100-seed campaign, 63.2\% of unconstrained Pareto-optimal candidates violate the pore-filling boundary, versus 39.3\% under the constraint; unconstrained candidates dominate the high-capacity region, where infeasibility concentrates.}\label{fig:pareto}
\end{figure}

\begin{figure}[t]
\centering
\includegraphics[width=0.9\textwidth]{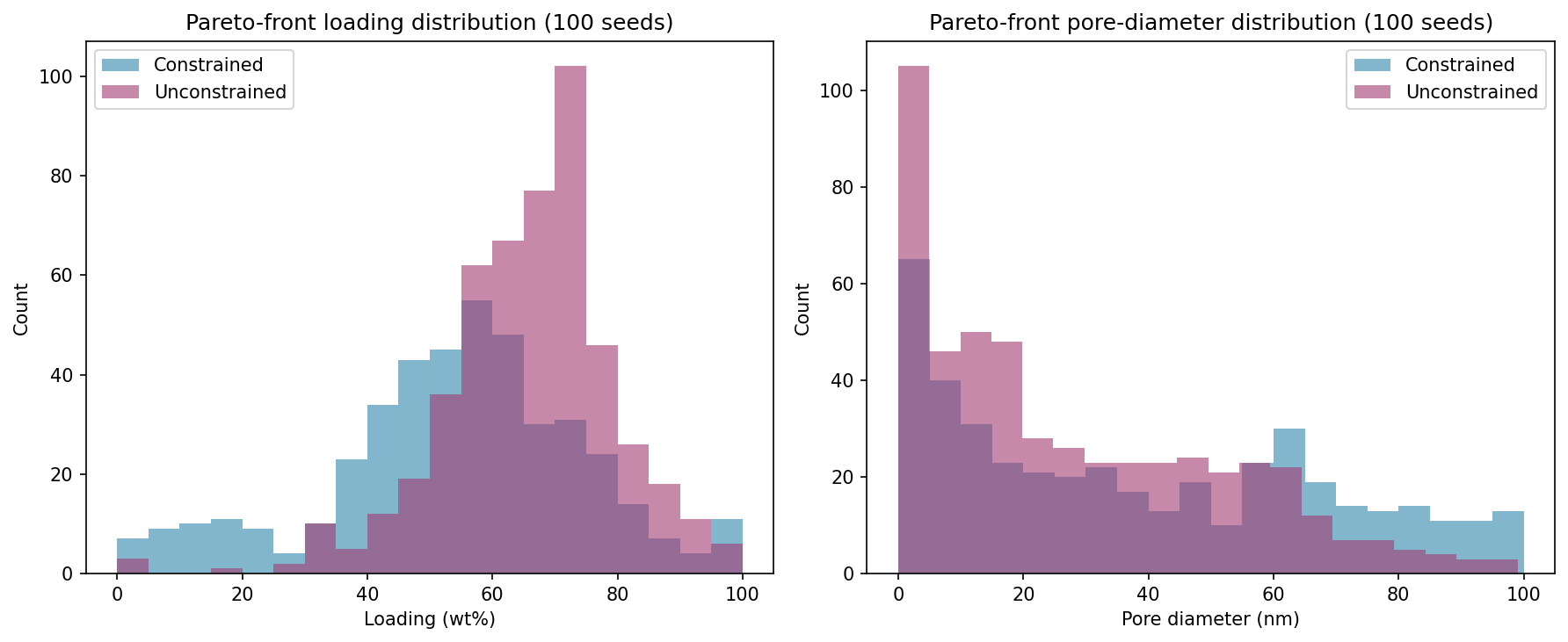}
\caption{\textbf{The constraint redirects the front toward a moderate-loading, larger-pore design window rather than merely truncating it at the boundary.} Marginal distributions of amine loading and pore diameter among Pareto-optimal candidates: the constraint shifts the front towards lower loadings (55.8 vs.\ 65.2 wt\%) and larger pores (36.8 vs.\ 27.7 nm).}\label{fig:marginals}
\end{figure}

Analysing the Pareto fronts constructed from the candidates each optimizer actually evaluated over 100 seeds (baseline history points excluded), \textbf{$63.2\pm30.9\%$ of the unconstrained front's candidates have predicted feasibility below zero, versus $39.3\pm28.0\%$ under the constraint}. Both fractions are higher on the merged benchmark than on the core library (unconstrained: 63.2\% here vs.\ 49.1\% on the core library; constrained: 39.3\% vs.\ 26.3\%): the newly merged high-capacity entries make the surrogate more optimistic in the high-loading, low-pore-volume region, increasing the pull toward geometrically inconsistent formulations for both optimizers---but the gap between them is preserved. The constraint steers the front toward lower amine loadings ($55.8$ vs.\ $65.2$ wt\%) and larger pores ($36.8$ vs.\ $27.7$ nm) (Figs.~\ref{fig:pareto}, \ref{fig:marginals}). The residual infeasible fraction under the constraint arises from the feasibility GP's predictive uncertainty in the smooth tanh transition zone ($0.9\lesssim\hat{\phi}\lesssim1.1$), where labels change sign over a narrow loading interval. Per-seed fronts are small (constrained: median 4 candidates, IQR 3--5; unconstrained: median 5, IQR 4--6), so per-seed infeasible fractions are coarsely quantized and pooled standard deviations are large; across the 100 seeds the per-seed infeasible fraction has median 33\% (IQR 20--50\%) under the constraint versus 67\% (IQR 40--88\%) without it. The direction and rough magnitude of the gap are stable; the absolute percentages should not be over-interpreted at this budget. The purpose of the smooth feasibility model is not to eliminate all boundary-touching proposals under surrogate uncertainty, but to reduce the optimizer's systematic drift toward the overfilled regime---which it does, roughly halving the infeasible fraction while retaining the capacity-seeking behaviour of the acquisition function. An oracle ablation sharpens this reading: replacing the feasibility-GP filter with the noiseless analytic constraint $c(x)$---computable here because the benchmark's ground truth is itself closed-form, but not in deployment, where a candidate's pore volume may be unmeasured---leaves final capacity statistically unchanged ($3.029\pm0.821$ vs.\ $2.964\pm0.728$ mmol\,g$^{-1}$; paired Wilcoxon $p=0.60$, 100 seeds) while reducing the front infeasible fraction from $39.3\%$ to $7.9\%$ ($p=6.0\times10^{-13}$; Supplementary Section S4). The headline infeasible fraction is therefore a conservative upper bound set by filter fidelity, not by the constraint.

\subsection{Baselines find capacity but not deployable formulations}\label{sec:baselines}

Nominal capacity alone does not discriminate the constrained optimizer from uninformed baselines (Table~\ref{tab:main}); deployability does. Constructing Pareto fronts from the candidates each baseline actually evaluated (100 seeds), the random-search front is $57.4\pm25.6\%$ infeasible (best capacity 4.592) and the single-objective-EI front $59.6\pm26.6\%$ infeasible (best capacity 5.577)---comparable to unconstrained MOBO (63.2\%) and far above the constrained front (39.3\%). The baselines find capacity but not deployable formulations: their apparent competitiveness in Table~\ref{tab:main} is carried by candidates whose amine volume exceeds the pore volume. We therefore report deployability alongside capacity throughout, and caution against evaluating formulation-optimization methods on surrogate capacity alone.

\subsection{Feasibility-blind metrics over-reward undeployable candidates}\label{sec:feashv}

Standard hypervolume (HV) credits every evaluated candidate, including geometrically inconsistent ones, and therefore systematically over-rewards methods that propose them. We define \textbf{feasible hypervolume (feasible-HV)} as $\mathrm{HV}_{\mathrm{feas}}=\mathrm{HV}(\{f(x_i):c(x_i)\geq0\})$, the HV computed only over candidates satisfying $c(x)\geq0$, using a shared reference point pooled across all methods and runs. Two properties matter. First, feasible-HV is evaluated over the full evaluated-candidate set of each run, not only the non-dominated front, so it prices the search budget spent on infeasible proposals rather than masking them post hoc. Second, the shared reference point keeps the metric comparable across methods, so the ratio $(\mathrm{HV}-\mathrm{feasible\text{-}HV})/\mathrm{HV}$ isolates each method's reliance on candidates outside the valid design space. This inflation, evaluated over the full 100-seed campaign, is \textbf{11.1\% for the constrained optimizer but 16.1\% for unconstrained MOBO, 13.9\% for random search, and 14.0\% for single-objective EI} (paired Wilcoxon on per-seed inflation, constrained vs.\ unconstrained: $p=2.1\times10^{-3}$): a systematic, statistically significant share of the feasibility-agnostic methods' apparent performance is an artifact of geometrically inconsistent candidates. Because rejected proposals are not replaced, the constrained arm evaluates fewer candidates per run ($42.9\pm9.9$ vs.\ 80); an equal-evaluation-budget control (30 seeds; Supplementary Section S5) reproduces the inflation gap at budget parity ($10.1\%$ vs.\ $18.5\%$ for unconstrained MOBO on the same seeds, paired Wilcoxon $p=0.003$), so no deployability conclusion depends on the budget asymmetry. The correction is consequential but not cosmetic-only: under standard HV the constrained optimizer trails unconstrained MOBO by 19\% (1.476 vs.\ 1.829), whereas under feasible-HV the gap narrows to 14\% (1.297 vs.\ 1.508); on the smaller core library the correction was strong enough to reverse the ranking entirely (constrained first, 1.007 vs.\ 0.964), so the residual gap on the merged benchmark is driven by the newly merged high-capacity entries rather than by a failure of the constraint. Single-objective EI attains the highest feasible-HV of all methods ($1.576\pm0.480$; Supplementary Table S2), which is expected rather than contradictory: feasible-HV prices the budget spent on infeasible proposals, but it cannot price objectives a method never optimized---EI concentrates its entire budget on capacity alone, 59.6\% of its Pareto-front candidates remain geometrically infeasible (Supplementary Table S1), and its feasible survivors carry no kinetics guarantee. The deployability claim therefore rests on the like-for-like multi-objective comparison, in which the constraint reduces infeasible front candidates from 63.2\% to 39.3\% while retaining 86\% of unconstrained MOBO's feasible-HV. We argue feasible-HV, or an equivalent feasibility-discounted metric, should accompany standard HV whenever candidate generation can leave the geometrically consistent design space (metric definitions, design choices, and the full results table: Supplementary Section S5, Table S2).

\subsection{Robustness to the feasibility threshold}\label{sec:tau}

Sweeping $\tau$ across seven quantiles of the benchmark feasibility-score distribution (corresponding to $\hat{\phi}$ limits between $\approx0.20$ and $\approx1.29$), the constrained capacity varies mildly and non-monotonically (2.409--3.050 mmol\,g$^{-1}$ across the whole sweep), with no systematic capacity gain from relaxing the threshold. The physical choice $\tau=0$ ($\hat{\phi}=1$) sits inside this stable plateau: the boundary behaves as a robust cutoff, not a fragile hyperparameter. Motivated by the amine-efficiency dip that already onsets at $0.9<\hat{\phi}\leq1.1$, an engineering margin $\tau=0.153$ ($\hat{\phi}\lesssim0.85$) preserves final capacity ($2.875\pm0.904$, 10 seeds, vs.\ $2.964\pm0.728$ over the 100-seed main campaign) while shifting the front further toward the deployable corner (mean loading 52.4 wt\%, mean pore diameter 38.0 nm) and reducing geometrically inconsistent ($\hat{\phi}>1$) evaluations from 25.9\% to 20.8\% of all evaluated candidates. The main conclusions are likewise insensitive to implementation choices: they hold under one-hot encoding of the categorical variables, under candidate generation restricted to the SBA-15 support family, and with a binary feasibility target in place of the smooth tanh score (Supplementary Section S6).

\subsection{An evolutionary baseline points to objective-awareness, not optimizer family}\label{sec:nsga2}

As an optimization-paradigm control, we ran NSGA-II \cite{deb2002,blank2020} under the identical surrogate, pinned DAC conditions, and evaluation budget (80 evaluations, 10 seeds). NSGA-II approaches the constrained optimizer's nominal capacity ($2.477\pm0.406$) but its Pareto front is $65.3\pm35.1\%$ infeasible at the highest mean loading of any method tested (72.3 wt\%)---confirming that the deployability gap is a property of objective-awareness, not of the optimizer family.

\subsection{Chance constraints and PV-aware filtering}\label{sec:chance}

Two refinements of the constraint itself prove less consequential than the constraint. A chance-constrained variant retaining only candidates with posterior $P(c(x)\geq0)\geq0.95$ under the feasibility GP (30 seeds) attains $2.635\pm0.599$ mmol\,g$^{-1}$---statistically indistinguishable from the mean-constrained optimizer ($p=0.15$)---and does not reduce the front's infeasible fraction (56.0\% vs.\ 43.2\% for the mean-constrained optimizer on the same 30 seeds), because the feasibility GP's variance is largest exactly where the tanh boundary flips sign, so a stricter probability threshold prunes borderline candidates without removing boundary errors. The informative fix is instead to propagate the uncertainty of the pore-volume lookup itself into the constraint: with first-order propagation of support- and family-level PV standard deviations, $\sigma_c=(1-\tanh^2(1-\hat{\phi}))\cdot\hat{\phi}\,\sigma_{PV}/PV$, and filtering at $P(c\geq0)\geq0.90$, the optimizer retains mean-constraint capacity (2.931) while reducing geometrically inconsistent evaluations to 26.4\%---a hyperparameter-free alternative to the engineering margin $\tau$ (20.8\%, Section~\ref{sec:tau}).

\subsection{Harmonized kinetics redirect the front}\label{sec:harmonized}

The kinetic objective compiled from the literature is apparatus-dominated (Section~\ref{sec:validation}b); we therefore treat kinetics throughout as a harmonized relative penalty indicator, not as an absolute optimization target. Replacing raw $1/t_{90}$ values with the fixed-effect predictions of the hierarchical model---i.e., optimizing against kinetics \emph{as they would be measured by a typical apparatus}---and rerunning the constrained MOBO (10 seeds) moves the Pareto front to markedly lower loadings and larger pores (mean loading 51.1 wt\%, mean pore diameter 49.5 nm, versus 55.8 wt\% and 36.8 nm against raw kinetics). Part of the high-loading recommendation in the raw analysis is therefore an apparatus artifact; harmonization removes it.

\subsection{A third objective: regeneration energy}\label{sec:energy}

\begin{figure}[t]
\centering
\includegraphics[width=0.8\textwidth]{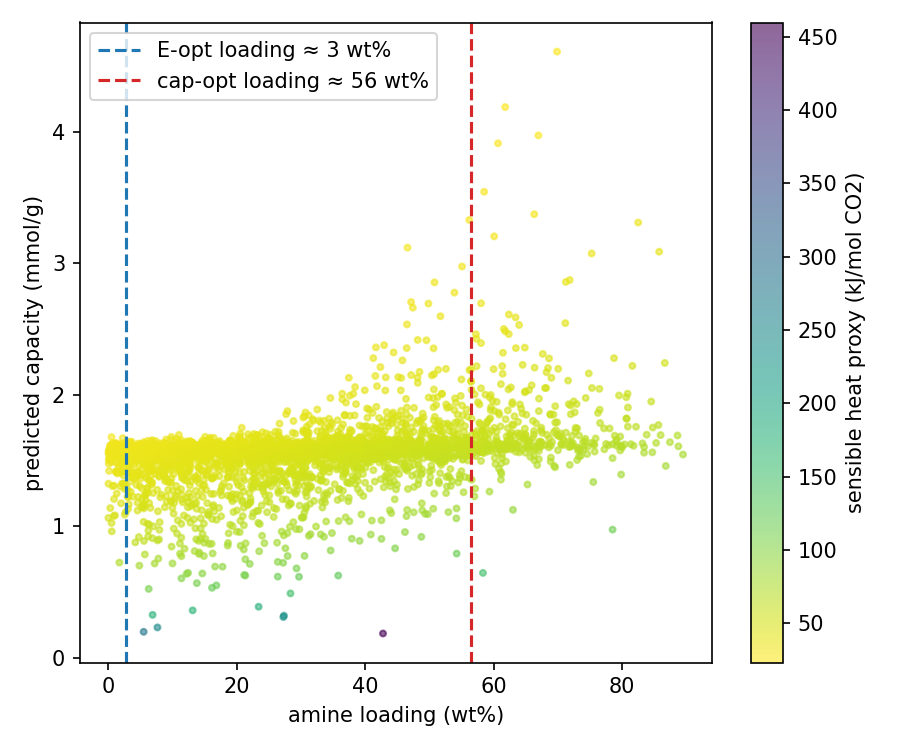}
\caption{Capacity--energy landscape over 5{,}000 surrogate-scored grid points. The capacity-optimal region ($\approx56$ wt\%) and energy-optimal region ($\approx3$ wt\%) are strongly misaligned; the compromise set concentrates near 47 wt\% loading and $\hat{\phi}\approx0.45$.}\label{fig:energy}
\end{figure}

Capacity and kinetics are not the only deployment-relevant objectives. The following is an illustrative third-objective stress test built on a sensible-heat proxy, not a process-level energy model. Approximating the sensible-heat contribution to regeneration energy as $E=c_{p,\mathrm{mix}}\,\Delta T/q$ kJ\,mol$^{-1}$ CO$_2$ (mixture heat capacity from support $c_p=0.8$ and amine $c_p=2.0$ J\,g$^{-1}$\,K$^{-1}$; $\Delta T=65$ K), and evaluating 5{,}000 surrogate-scored grid points, the capacity-optimal region sits at $\approx56$ wt\% loading and $\hat{\phi}\approx0.5$ ($E\approx46$ kJ\,mol$^{-1}$), while the energy-optimal region collapses to very low loadings ($\approx3$ wt\%, capacity 1.59 mmol\,g$^{-1}$). The latter corner is a surrogate extrapolation---its implied amine efficiency exceeds anything in the benchmark---and should be read as a direction, not a prediction. The capacity--energy compromise set (68 of 5{,}000 points) concentrates at \textbf{$\approx47$ wt\% loading, $\hat{\phi}\approx0.45$, capacity 2.39 mmol\,g$^{-1}$, $E\approx39$ kJ\,mol$^{-1}$} (dimension-wise medians over the 68 set members, not a single grid point; Fig.~\ref{fig:energy}; per-point values and a self-consistency note: Supplementary Section S4); across five independent grid seeds the compromise remains within 46.8--51.2 wt\% and $\hat{\phi}=0.42$--0.49.

Three further analyses are reported in the Supplementary Information and drawn upon where relevant below: an interpretable decision-tree distillation of the surrogate's behaviour in the DAC window (Supplementary Section S1); out-of-sample recommendation checks---out-of-time, leave-one-publication-out, and out-of-benchmark---reported at region level with null results included (Supplementary Section S2); and two boundary audits, on support featurization and on humidity response (Supplementary Section S3).

\section{Discussion}\label{sec:discussion}

We encoded the pore-filling limit as an outcome constraint on multi-objective Bayesian optimization over DAC sorbent formulations and used it as a diagnostic instrument. The constraint lowers nominal surrogate-predicted capacity, and much of the unconstrained optimizer's apparent advantage is carried by candidates whose amine volume exceeds the pore volume of their support: a deployability-aware metric (feasible hypervolume) shows that 13.9--16.1\% of feasibility-agnostic methods' hypervolume is an artifact of geometrically inconsistent candidates, versus 11.1\% for the constrained optimizer. The boundary is physically derived rather than fit to the optimization target: its sign is learnable across laboratories (AUC $=0.93$), the amine-efficiency dip in the transition zone provides partial experimental corroboration, and an external check against directly measured pore filling agrees within 0.4--2.1 percentage points.

Three surrogate-based analyses converge on the same candidate design window. The constrained Pareto front centres near 56 wt\% loading; replacing raw kinetics with apparatus-harmonized kinetics moves it to 51 wt\%; and the capacity--energy compromise set concentrates near 47 wt\% at $\hat{\phi}\approx0.45$. A formulation designer reading this literature---or an autonomous system mining it---should preferentially explore approximately \textbf{45--55 wt\% oligoamine loading at pore filling $\hat{\phi}\approx0.4$--0.5}, well inside the geometric boundary, rather than the nominal capacity maximum that unconstrained optimization will reliably propose and that the literature's own capacity trend already penalizes beyond $\hat{\phi}>1$. The wt\% range assumes the high pore volumes (PV $\approx2$ cm$^3$\,g$^{-1}$) of the silicas dominating this window; on leaner supports the same $\hat{\phi}$ window maps to lower loadings, so $\hat{\phi}$ is the portable coordinate. The prominence of oligoamines in this window is chemically plausible, given the cooperative CO$_2$ insertion documented between neighbouring amine sites \cite{mcdonald2015}.

\textbf{Meta-scientific lessons.} This study doubles as an audit of the DAC sorbent literature. Three observations generalize beyond our optimizer. First, \emph{interoperability is the binding constraint of literature mining}: of 68 independently curated candidate entries, only 27 could be merged---grafted sorbents report no impregnation loading, breakthrough protocols report $t_5$/$t_{50}$ rather than batch $t_{90}$, and LDF constants are not convertible across apparatus classes. Reporting practice, not data quality, excluded the majority. Second, \emph{apparatus effects dominate kinetics} (77.8\% of variance between publications); pooled kinetic analyses are uninformative until harmonized hierarchically, after which the physically expected kinetic cost of pore filling emerges ($p=0.002$, retained under a random-slope refit at $p=0.010$ and after excluding all $t_{50}$-converted records at $p=0.012$). Third, \emph{recent literature shows a shift toward lower pore-filling ratios}: median pore filling of new formulations declined from 0.849 (2021--22) to 0.393 (2025+) (Supplementary Section S2). We therefore report our weak and null results on equal footing with the strong ones: chance-constrained filtering without pore-volume uncertainty propagation does not outperform mean constraints; leave-one-publication-out checks aggregate to parity with random outside well-sampled chemistry (Supplementary Section S2); and no featurization scheme we tested extrapolates to unseen support classes (Supplementary Section S3). These boundaries, not only the headline numbers, define where literature-mined optimization can presently be trusted.

\textbf{Limitations.} (i) This work is surrogate-based simulation, not a wet-lab closed loop. (ii) The surrogate is trained on condition-heterogeneous data dominated by flue-gas-scale measurements; only a minority of entries are measured at DAC conditions, so capacity predictions at the pinned DAC point are extrapolations moderated by the GP, and absolute values should be read as surrogate estimates. (iii) The compiled kinetics are apparatus-dominated; only after hierarchical harmonization does the expected kinetic cost of pore filling emerge, and standardized kinetic reporting would materially strengthen any literature-mined kinetics benchmark. (iv) For $\hat{\phi}\gtrsim1.1$ the geometric assumptions break down (swelling, external amine, underestimated pore volumes), bounding the constraint's validity. (v) The constraint inherits a single median pore volume per support class; the PV-aware variant (Section~\ref{sec:chance}) propagates the associated uncertainty but cannot remove it. (vi) Part of the kinetics values are converted from $t_{50}$ under a first-order assumption. (vii) Prospective and retrospective checks are limited by publicly confirmable metadata and by the chemical skew of the literature toward SBA-15-type silicas; hit rates are reported at region level, not pointwise. The constrained/unconstrained deployability gap, however, persists within the SBA-15 family alone (Supplementary Section S6), so it is not a support-mixing artifact. (viii) Humid-condition behaviour is audited but not modelled.

\section{Methods}\label{sec:methods}

\subsection{Problem formulation}

Let $x=(s,a_1,a_2,L,B,D,T,c,h,F)$ denote a formulation--condition pair, where $s$ is the support class, $a_1,a_2$ the (optionally blended) amine classes, $L$ the amine loading (wt\%), $B$ the BET area (m$^2$\,g$^{-1}$), $D$ the average pore diameter (nm), and $(T,c,h,F)$ the operating conditions (temperature, CO$_2$ concentration, relative humidity, flow rate). Objectives are equilibrium capacity $f_{\mathrm{cap}}(x)$ (mmol\,g$^{-1}$, maximize) and kinetics $f_{\mathrm{kin}}(x)=1/t_{90}$ (min$^{-1}$, maximize). Operating conditions are model features, not filters; during candidate generation they are pinned to the DAC deployment point ($T=25\,^\circ$C, $c=0.04$ vol\%, $h=0$, $F=100$ mL\,min$^{-1}$).

\subsection{Feasibility constraint}

For amine mass fraction $w=L/100$ and support pore volume $PV$ (cm$^3$\,g$^{-1}$), the pore-filling ratio is
\begin{equation}\label{eq:phi}
\hat{\phi}=\frac{w/\rho_{\mathrm{amine}}}{PV\,(1-w)},
\end{equation}
the volume of impregnated amine (density $\rho_{\mathrm{amine}}$ tabulated per amine class, 0.90--1.10 g\,cm$^{-3}$) divided by the dry-support pore volume. $\hat{\phi}\geq1$ is geometrically inconsistent with pore-confined impregnation. Strictly, this does not mean the formulation cannot be synthesized; it means the impregnated amine can no longer be accommodated within the measured pore volume, so the pore-confined picture on which both the design rule and the surrogate rest ceases to apply (swelling, external amine films, or underestimated pore volumes; see Limitations). We map $\hat{\phi}$ to a smooth score
\begin{equation}\label{eq:c}
c(x)=\tanh(1-\hat{\phi})-\tau,
\end{equation}
with physical threshold $\tau=0$ unless otherwise swept; the tanh transform retains gradient information in the over-filled regime. Pore volume is not an optimization variable; candidates inherit the median pore volume of their support class, with support-level and family-level standard deviations retained for uncertainty propagation (below).

\subsection{Multi-objective Bayesian optimization}

Two independent SingleTaskGP models (GPyTorch \cite{gardner2018}) model $f_{\mathrm{cap}}$ and $f_{\mathrm{kin}}$; a third models $c(x)$. Categorical variables are ordinal-encoded (robustness to one-hot encoding: Supplementary Section S6); hyperparameters are re-estimated by maximum marginal likelihood after each batch. We use qLogEHVI \cite{daulton2020,ament2023,balandat2020}, the numerically stabilized variant of qEHVI, with the constraint entering as feasibility weighting in the acquisition function rather than as a penalty on the objectives \cite{gelbart2014,letham2019}. Each run starts from 10 randomly sampled history points and proceeds for 20 iterations; each iteration proposes up to 4 candidates by joint acquisition optimization ($q=4$, 2 restarts, 32 raw samples), evaluated against the literature-calibrated surrogate and appended before refitting. Under the constraint, batches may contain fewer than 4 candidates; rejected proposals are not replaced, so the constrained optimizer never sees infeasible regions as data. This costs the constrained arm about half the true evaluation budget; an equal-evaluation-budget control verifies that no conclusion depends on the asymmetry (Supplementary Section S5). Results aggregate 100 seeds unless stated.

\subsection{Baselines and metrics}

Baselines: random search; single-objective EI on capacity; NSGA-II \cite{deb2002} via pymoo \cite{blank2020} (population 16, 5 generations, SBX crossover and polynomial mutation within decision bounds). Unless stated otherwise, all analyses use the merged 618-entry benchmark. Standard hypervolume uses the pooling-based reference point of Zitzler--Thiele-type \cite{zitzler1999,while2006}; feasible-HV restricts the candidate set to $c(x)\geq0$ before HV computation, sharing the reference point across methods. Chance constraints filter candidates by posterior $P(c(x)\geq0)$; the PV-aware variant propagates pore-volume uncertainty as $\sigma_c=(1-\tanh^2(1-\hat{\phi}))\cdot\hat{\phi}\,\sigma_{PV}/PV$, with $\sigma_{PV}$ the largest of support-level, family-level, or a 10\% floor.

\subsection{Hierarchical kinetic harmonization}

A linear mixed-effects model $\log_{10}(1/t_{90})\sim T+\log_{10}c_{\mathrm{CO_2}}+L+\hat{\phi}+(1\mid\mathrm{publication})$ (statsmodels \cite{seabold2010}) partitions apparatus variance (ICC) from fixed physicochemical effects; the fixed-effect predictor defines the harmonized kinetic target used in Section~\ref{sec:harmonized}.

\subsection{Regeneration-energy proxy}

$E=c_{p,\mathrm{mix}}\Delta T/q$ kJ\,mol$^{-1}$ CO$_2$, with $c_{p,\mathrm{mix}}=(1-w)\,c_{p,\mathrm{support}}+w\,c_{p,\mathrm{amine}}$, $c_{p,\mathrm{support}}=0.8$, $c_{p,\mathrm{amine}}=2.0$ J\,g$^{-1}$\,K$^{-1}$, $\Delta T=65$ K, and $q$ the (surrogate-predicted) equilibrium capacity in mmol\,g$^{-1}$ (J\,g$^{-1}$ over mmol\,g$^{-1}$ gives J\,mmol$^{-1}=$ kJ\,mol$^{-1}$; sensible heat only---reaction enthalpy, heat losses, and process-level duties excluded).

\subsection{Statistics}

Seed-level comparisons use two-sided Wilcoxon signed-rank tests; group medians use Mann--Whitney $U$; correlations are Spearman's $\rho$. Uncertainty intervals are standard deviations across seeds. Featurization audits use scikit-learn \cite{pedregosa2011} GP regressors under leave-one-publication-out and leave-one-support-out protocols. During the preparation of this work, the authors used Kimi (Moonshot AI) and GPT (OpenAI) for language editing, manuscript restructuring, and code review; the authors reviewed and edited all outputs and take full responsibility for the content.

\backmatter

\bmhead{Supplementary information}
Supplementary Information is available for this paper: S1 interpretable design rules; S2 out-of-sample checks (out-of-time, LOPO recommendation, out-of-benchmark evidence table); S3 boundary audits (support featurization, humidity pair list); S4 additional robustness analyses (100-seed seed-extension, pore-volume sensitivity, energy-proxy sensitivity); S5 feasible-HV metric definitions and design choices; S6 encoding, support-family, and constraint-form robustness; amine density table; support median pore-volume table; stage-0 merge disposition (68$\to$27, per-entry reasons); LOPO model details; NSGA-II settings; mixed-model full output.

\bmhead{Acknowledgements}
The research of Y.L. was supported by the National Natural Science Foundation of China (no. 12001306). This work was supported by the National Science Fund for Distinguished Young Scholars (52225407); the National Natural Science Foundation of China (52522412); the Key Program of Joint Funds of the National Natural Science Foundation of China (U24A2046); the Guangdong Major Project of Basic and Applied Basic Research (2023B0303000024); and the Shenzhen Science and Technology Program (KCXFZ20240903093600001 and KCXFZ20230731093902006). 

\textbf{Use of AI-assisted tools.} During the preparation of this manuscript, AI-assisted tools were used to support language editing, manuscript restructuring, and code review. All AI-assisted outputs were critically reviewed and verified by the authors, who take full responsibility for the accuracy, integrity, and content of the manuscript.

\bmhead{Author contributions}
Y.L.: conceptualization, methodology, software, formal analysis, investigation, data curation, visualization, writing---original draft. X.S.: validation, writing---review \& editing. Z.Z.: supervision, project administration, funding acquisition, writing---review \& editing. All authors read and approved the final manuscript.

\bmhead{Data availability}
The curated benchmark and associated analysis outputs generated in this study are available in a review-accessible repository provided with this submission. The benchmark's 591-entry core is derived from the public dataset of Wang \emph{et al.} \cite{wang2025scidata} (CC BY 4.0); the filtering and harmonization steps and the 27 independently curated additions are documented with per-field source annotations and per-row provenance flags in the repository. A persistent DOI will be minted upon publication; the arXiv version of this article will be updated with the repository link once it is live.

\bmhead{Code availability}
Custom code supporting the findings of this study is not publicly available at this stage; the code required to reproduce the benchmark construction, optimization experiments, and analysis results reported in this study is available to editors and reviewers during peer review, and from the corresponding author on reasonable request.

\bmhead{Competing interests}
The authors declare no competing interests.

\clearpage
\setcounter{section}{0}
\setcounter{figure}{0}
\setcounter{table}{0}
\setcounter{equation}{0}
\setcounter{page}{1}
\renewcommand{\thesection}{S\arabic{section}}
\renewcommand{\thefigure}{S\arabic{figure}}
\renewcommand{\thetable}{S\arabic{table}}
\renewcommand{\theequation}{S\arabic{equation}}

\begin{center}
{\LARGE\bfseries Supplementary Information\par}
\vspace{0.8em}
{\large\bfseries When High Capacity Becomes Undeployable: Pore-Filling Constraints for Multi-Objective Optimization of Direct Air Capture Sorbents\par}
\vspace{0.8em}
{Yujie Liu$^{1,*}$, Xuehua Shen$^{1}$, Zuotai Zhang$^{1,*}$\par}
\vspace{0.3em}
{\small $^{1}$Southern University of Science and Technology, Shenzhen, 518055 Guangdong, China. $^{*}$e-mail: liuyj@sustech.edu.cn; zhangzt@sustech.edu.cn\par}
\end{center}
\vspace{0.5em}
\noindent\textbf{Data provenance note.} The 591-entry core library analysed throughout this work is derived from the public dataset of Wang et al., \emph{Scientific Data} \textbf{12}, 724 (2025), DOI: 10.1038/s41597-025-05037-1 (CC BY 4.0), filtered and harmonized as described in the main text; the remaining 27 entries were curated independently for this study and carry per-row provenance flags in the accompanying repository.

\section{Interpretable design rules}\label{sec:s1}

To distil the surrogate's behaviour into an actionable rule, we trained a depth-4 decision tree on the DAC-window subset ($n=146$), labelling entries with capacity $\geq$ the upper quartile (1.783 mmol\,g$^{-1}$) and $\hat{\phi}\leq1$ as positive (28 entries). The resulting rule (92.5\% training accuracy, versus an 80.8\% majority-class baseline; balanced accuracy 88.5\%) is a single sentence: \emph{oligoamine-class amines at loading $>32$ wt\% with $\hat{\phi}\leq0.93$ yield high-performance formulations}. Feature importances---amine class 0.42, loading 0.25, $\hat{\phi}$ 0.22---indicate that the pore-filling ratio is not merely a post-hoc filter but itself a top-three predictor of high performance. We stress that this rule is an interpretive distillation of the surrogate's behaviour on the benchmark, not an independently validated design law. It is, however, not a training-set artifact: under publication-grouped 10-fold cross-validation (no test record shares a publication with its fold's training set), the same depth-4 tree retains 87.4\% out-of-fold accuracy (balanced accuracy 79.6\%, positive-class F1 0.679; depths 2--5 give 84.9--87.4\%), and the amine-class/loading/$\hat{\phi}$ importance ordering is stable across folds.

\section{Out-of-sample checks: where recommendations can and cannot be trusted}\label{sec:s2}

\begin{figure}[t]
\centering
\includegraphics[width=0.7\textwidth]{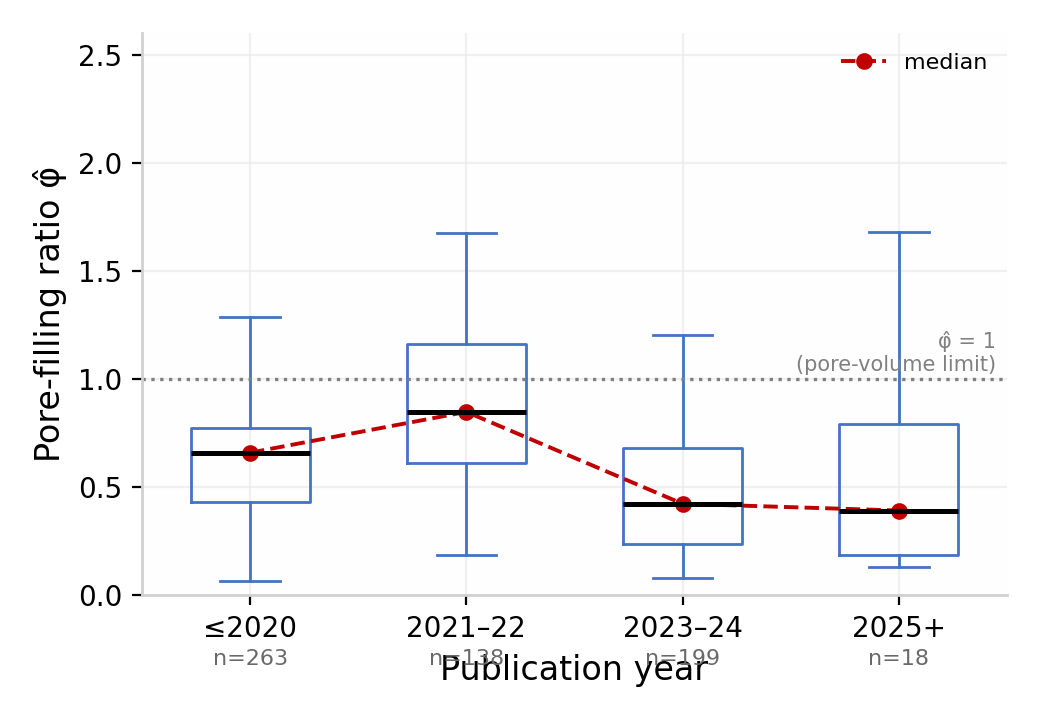}
\caption{Median pore-filling ratio $\hat{\phi}$ of newly published formulations by year bin ($n=618$). The field is converging toward the conservative interior of the feasible region ($\hat{\phi}<1$), not toward the geometric boundary.}\label{fig:s_year}
\end{figure}

We frame the following as boundary checks rather than validation: their statistical power is limited by the small size of the post-2022 literature and by its chemical skew, and we report them accordingly---at region level, with null results included. \textbf{Out-of-time split.} Training the pipeline on publications dated $\leq2022$ (30 publications) and testing recommendations against 2023+ publications (13), the constrained optimizer's per-seed hit rate (same support family, same amine class, $|\Delta$loading$|\leq10$ wt\%) is 0.13--0.33 versus 0.12--0.25 for random search. More informative than the hit rate is the drift it reveals: median $\hat{\phi}$ of newly published formulations declined from 0.849 (2021--22) to 0.424 (2023--24) and 0.393 (2025+) (Fig.~\ref{fig:s_year}). The field is converging toward the conservative interior of the feasible region, not toward the geometric boundary---the constraint makes explicit, and optimizable, a trend the community was already following implicitly. \textbf{Leave-one-publication-out (LOPO) recommendation.} Holding out each of 18 DAC-window publications and asking whether the optimizer's top-5 capacity recommendations recover the held-out formulations, the aggregate hit rate is on par with random (32\% vs.\ 30\%)---but this average conceals structure: on publications that themselves contain frontier formulations (window capacity $\geq$ upper quartile), the optimizer leads (42\% vs.\ 38\%), and on the most recent systematic study of impregnated SBA-15 sorbents it recovers all held-out formulations (100\% vs.\ 40\%). Retrospective recommendation is reliable precisely where the literature has sampled densely (SBA-15 $\times$ oligoamines---the field's home turf), and unreliable for chemically distinct one-off studies, consistent with the featurization audit of Section~\ref{sec:s3}. \textbf{Out-of-benchmark external literature check.} On five recent (2023--2025) reports outside the benchmark with confirmable metadata, all formulations with complete pore-volume data fall inside the predicted deployable region, and our first-principles $\hat{\phi}$ agrees with directly measured (N$_2$ physisorption) pore filling within 0.4--2.1 percentage points. Within the benchmark's DAC window, median capacity peaks in the 0.9--1.0 bin (1.54 mmol\,g$^{-1}$) and drops beyond (1.23 mmol\,g$^{-1}$ at $\hat{\phi}>1$): the region the constraint excludes is also where the literature's own capacity trend reverses. The intermediate bin medians are non-monotone and confounded by support-family composition across bins (low-$\hat{\phi}$ bins are dominated by low-pore-volume hosts), so we cite only the peak-versus-beyond contrast. Per-report DOIs, formulations, pore volumes, computed $\hat{\phi}$, measured pore-filling estimates, and errors are tabulated in the released dataset (\texttt{oot\_validation.csv}); of the five reports, three provide complete pore-volume data, one reports only a pore-volume range (evaluated as an interval), and one grafted-amine study reports no impregnation loading, so $\hat{\phi}$ is undefined---the same exclusion logic as the stage-0 curation described in the main text.

\section{Boundary audits: support featurization and humidity}\label{sec:s3}

\begin{figure}[t]
\centering
\includegraphics[width=0.8\textwidth]{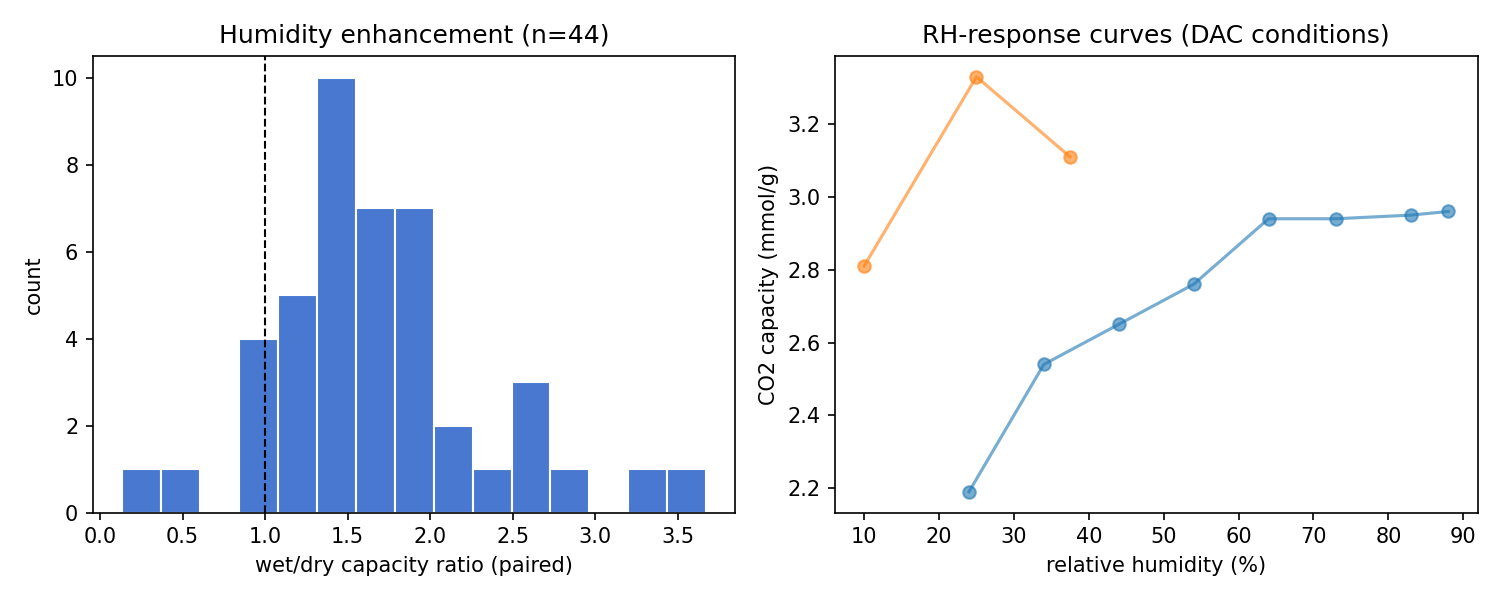}
\caption{Humidity response in the raw literature extraction: 44 matched wet/dry pairs give a median capacity enhancement of 1.594$\times$ (86\% enhanced), with a plateau onset near 44.5\% RH.}\label{fig:s_humidity}
\end{figure}

Two audits bound where the present models can be trusted. First, replacing support identity with texture-based descriptors (three featurization schemes) leaves within-publication prediction broadly comparable (LOPO $R^2\approx0.51$--$0.53$, versus $0.57$--$0.60$ with support identity---a modest but real drop), but no scheme extrapolates to unseen support classes (leave-one-support-out median $R^2=-0.68$ to $-0.79$): generic texture recovers much---but not all---of the predictive content of support identity, and cross-support zero-shot prediction is currently unreliable. Second, humidity---excluded from the main (dry) benchmark---is systematically beneficial in the raw extraction: across 44 matched wet/dry pairs, the wet-to-dry capacity ratio has median 1.594 (IQR 1.30--1.98; 86\% enhanced), with an enhancement plateau onset near 44.5\% relative humidity (Fig.~\ref{fig:s_humidity}), consistent with water-promoted carbamate chemistry. (The raw extraction contains 45 humid DAC records; 44 have a dry-condition partner from the same study---same support, amine, loading within 2~wt\%, temperature within 3~$^\circ$C---and form the paired ratios, while the remaining record enters only the RH-response curves.) Dry-condition conclusions are therefore likely conservative for humid deployment, but humid-condition kinetics and stability data remain too sparse to model.

\section{Additional robustness analyses}\label{sec:s4}

\textbf{Seed extension of the main campaign.} The main-text comparison (Table~1 of the main text) aggregates 100 independent seeds per method. Table~\ref{tab:s_d1} reports the full statistics, including hypervolume under the shared reference point and per-seed Pareto-front size and infeasible-fraction summaries. The direction and significance of every main-text comparison already hold at 30 seeds; the extension to 100 seeds tightens the Wilcoxon $p$-values from $<10^{-4}$ to $\leq10^{-11}$ for both Bayesian baselines and preserves the null result against random search. Effect sizes accompany every $p$-value (d10\_effect\_sizes\_v5.json; constrained minus comparator, so negative values mean lower constrained nominal capacity): versus unconstrained MOBO, rank-biserial correlation $-0.79$ with median paired difference $-0.564$ [$-0.809,-0.419$]; versus single-objective EI, $-0.82$ with $-0.905$ [$-1.096,-0.724$]---both large effects; versus random search, $-0.03$ with $-0.025$ [$-0.255,0.194$]---a null effect consistent with the $p$-value.

\begin{table}[h]
\centering
\caption{Main-campaign statistics over 100 seeds per method (20 iterations, batch $\leq4$; shared hypervolume reference point $[-0.483, -0.196]$). Wilcoxon signed-rank tests pair each method against constrained MOBO by seed.}\label{tab:s_d1}
\footnotesize
\setlength{\tabcolsep}{2pt}
\begin{tabular}{lccccc}
\toprule
Method & Final capacity & HV (final) & Front size (IQR) & Front infeasible & $p$ vs.\ constrained \\
\midrule
Constrained MOBO & $2.964\pm0.728$ & $1.476\pm0.475$ & 4 (3--5) & $39.3\pm28.0\%$ & -- \\
Unconstrained MOBO & $3.639\pm0.815$ & $1.829\pm0.564$ & 5 (4--6) & $63.2\pm30.9\%$ & $1.0\times10^{-11}$ \\
Single-objective EI & $3.826\pm0.823$ & $1.827\pm0.419$ & 4 (3--6) & $59.6\pm26.6\%$ & $1.3\times10^{-12}$ \\
Random search & $2.944\pm0.533$ & $1.397\pm0.227$ & 4.5 (3--6) & $57.4\pm25.6\%$ & $0.78$ (n.s.) \\
\bottomrule
\end{tabular}
\end{table}

\textbf{Pore-volume lookup sensitivity.} Pore volumes in the benchmark are predominantly measured, not imputed: 609 of 618 rows carry a measured PV, with only 9 rows assigned a class-median value. As a counterfactual stress test, replacing \emph{all} PV values by class medians leaves the feasibility label unchanged for 93.0\% of all rows and 96.6\% of DAC-window rows (Cohen's $\kappa=0.736$ and $0.864$, respectively); perturbing each PV by $\pm20\%$ moves the benchmark-wide feasible fraction only within 0.70--0.89 (baseline 0.845, reproduced exactly). The pore-filling labels driving the constraint are therefore robust to the PV lookup.

\textbf{Energy-proxy sensitivity.} The regeneration-energy analysis of the main text uses a sensible-heat proxy $E=c_{p,\mathrm{mix}}\Delta T/q$. Across 108 combinations of heat-capacity mixing rules ($3\times3$), regeneration temperature swings ($\Delta T$, 3 levels), and adsorption-enthalpy offsets ($E_{\mathrm{ads}}$, 4 levels), the compromise loading is stable and stays inside the main-text 45--55 wt\% window in 100\% of combinations (median 48.2 wt\%, range 47.1--49.5), while the compromise $\hat{\phi}$ spans 0.435--0.519 (88.9\% of combinations inside the 0.4--0.5 band, the remainder marginally above it). The adsorption-enthalpy offset translates absolute $E$ values without changing the ranking or the compromise location. Across five independent grid seeds of the main-text configuration, the compromise remains within 46.8--51.2 wt\%, $\hat{\phi}=0.42$--$0.49$, and $E=38.6$--$39.7$ kJ\,mol$^{-1}$. \textbf{Self-consistency of the reported compromise statistics.} The main-text compromise numbers are dimension-wise medians over the 68 set members, not the coordinates of a single grid point, so the proxy need not return the median $E$ when evaluated at the median loading and capacity: at 47.1 wt\% and 2.39 mmol\,g$^{-1}$ it gives $E\approx37$ kJ\,mol$^{-1}$, inside the compromise set's interquartile range of 35--41 kJ\,mol$^{-1}$, while the reported $E\approx39$ kJ\,mol$^{-1}$ is the median of the per-point proxy values. The per-point compromise-set values underlying both statistics are provided with the released analysis outputs (Data availability).

\textbf{CO$_2$-concentration-stratified surrogate error.} Because 57.1\% of benchmark records are measured at 100 vol\% CO$_2$ while every optimization is pinned to the DAC deployment point (0.04 vol\%), the surrogate's behaviour inside the DAC window deserves direct scrutiny. A 10-fold cross-validation of the benchmark surrogate (Table~\ref{tab:s_co2strata}) shows no extrapolation penalty for capacity: out-of-fold error in the DAC window ($n=173$; MAE $=0.29$ mmol\,g$^{-1}$, $R^2=0.79$) is comparable to the flue-gas mid-concentration and pure-CO$_2$ bands ($R^2=0.90$ and $0.86$) and is the lowest of the three bands in absolute terms. Kinetics is the noisier objective in every band---within-band $R^2$ spans $-1.65$ (DAC window, where the residual variance, s.d.\ $=0.027$ min$^{-1}$, is small and dominated by inter-laboratory scatter) to $0.81$ (mid band)---which is exactly the behaviour expected from the main-text variance decomposition (77.8\% publication-level variance, ICC $=0.778$) and motivates treating kinetics as a harmonized relative penalty indicator rather than an absolute optimization target throughout.

\begin{table}[h]
\centering
\caption{Ten-fold cross-validated error of the benchmark surrogate, stratified by CO$_2$ concentration band. Capacity in mmol\,g$^{-1}$; kinetics as $1/t_{90}$ in min$^{-1}$. $R^2$ is computed within each band. Note that the stratification bands are defined by CO$_2$ concentration alone: the DAC band ($<1$ vol\%, $n=173$) is broader than the temperature-restricted DAC window (20--35\,$^\circ$C, $\leq600$ ppm, $n=146$) used for the design rules of Section~\ref{sec:s1}.}\label{tab:s_co2strata}
\begin{tabular}{lcccccc}
\toprule
Band & $n$ & \multicolumn{2}{c}{Capacity} & \multicolumn{2}{c}{Kinetics} \\
\cmidrule(lr){3-4}\cmidrule(lr){5-6}
 & & MAE & $R^2$ & MAE & $R^2$ \\
\midrule
DAC window ($<1$ vol\%) & 173 & 0.29 & 0.79 & 0.027 & $-1.65$ \\
Mid concentration (1--100 vol\%) & 92 & 0.32 & 0.90 & 0.048 & 0.81 \\
Pure CO$_2$ (100 vol\%) & 353 & 0.34 & 0.86 & 0.169 & 0.48 \\
\bottomrule
\end{tabular}
\end{table}

\textbf{Publication-grouped cross-validation.} Random $k$-fold splits let records from one publication appear in both training and test folds; because records from one publication share apparatus, protocols, and material families, random CV measures interpolation rather than cross-publication generalization. Repeating the evaluation under GroupKFold by DOI (10 folds; no test record shares a publication with its fold's training set; Table~\ref{tab:s_groupedcv}), the decay is the expected one: capacity $R^2$ within the DAC band drops from 0.79 (random split) to 0.33, and the absolute error rises from 0.29 to 0.59 mmol\,g$^{-1}$---still the lowest of the three bands, and below the per-band standard deviation of measured capacities (0.91 mmol\,g$^{-1}$) in every band. Two conclusions survive the stricter split: rankings \emph{within} publication families remain informative, and the feasibility GP on which the constraint relies degrades far less than the objectives (Table~\ref{tab:s_feascls}). We therefore state all headline optimization conclusions as optimizer behaviour on the literature-calibrated surrogate---which is what a screening instrument is---and not as predictions of untested materials' absolute performance. Kinetics does not survive the grouped split at all: in the DAC band the grouped kinetic surrogate attains $R^2=-13.2$ (Table~\ref{tab:s_groupedcv}), far worse than the random-$k$-fold analogue ($-1.65$, Table~\ref{tab:s_co2strata}; the two validations are labelled separately throughout). All kinetic conclusions in this work are therefore association-level evidence from the harmonization analysis below and a relative penalty indicator---never predictions of unseen publications' absolute kinetics.

\textbf{Robustness of the hierarchical kinetic analysis.} The main-text mixed-effects model ($\log_{10}(1/t_{90})\sim T+\log_{10}c_{\mathrm{CO_2}}+L+\hat{\phi}+(1\mid\mathrm{publication})$) is robust to effect heterogeneity and collinearity (exp\_16\_mixedlm\_sensitivity.json). A random-slope specification retaining a per-publication $\hat{\phi}$ slope converges cleanly and leaves the fixed effect significant and negative ($\beta=-0.420$, 95\% CI $[-0.737,-0.102]$, $p=0.0096$; per-publication slope variance $0.235$), so the $\hat{\phi}$--kinetics association is not driven by pooling publications with heterogeneous slopes. Collinearity is modest: $\hat{\phi}$ and loading correlate structurally (Pearson $0.56$, Spearman $0.73$---expected, as $\hat{\phi}$ is a closed-form function of loading) yet the fixed-design variance inflation factors are $1.5$ ($\hat{\phi}$) and $1.5$ (loading), well below conventional concern thresholds. The intercept-only refit reproduces the frozen main-text values exactly (ICC $0.778$, $\beta=-0.184$, $p=0.0024$). The association also survives removal of every $t_{50}$-converted record: refitting both specifications on the 507 directly reported $t_{90}$ values (39 publications) keeps the $\hat{\phi}$ fixed effect negative and significant (random intercept: $\beta=-0.160$, 95\% CI $[-0.285,-0.035]$, $p=0.012$; random slope: $\beta=-0.259$, 95\% CI $[-0.510,-0.007]$, $p=0.044$; exp\_18\_t50\_exclusion.json), so the kinetic penalty is not an artefact of the first-order $t_{50}\!\rightarrow\!t_{90}$ conversion. The within-publication corroboration quoted in the main text is documented per publication in Table~\ref{tab:s_withinpaper}. We report this analysis as evidence of an association between pore filling and harmonized kinetics, not as a transferable kinetic prediction model.

\begin{table}[h]
\centering
\caption{Within-publication Spearman correlations between pore filling $\hat{\phi}$ and harmonized log-rate $\log_{10}(1/t_{90})$, for the 19 publications with $\geq8$ records spanning a $\hat{\phi}$ range (kinetics\_harmonization.json). 16 of 19 correlations are negative (median $-0.279$); $^{*}$marks the 7 individually significant at $p<0.05$, one of which is positive.}\label{tab:s_withinpaper}
\footnotesize
\setlength{\tabcolsep}{4pt}
\begin{tabular}{llcc}
\toprule
DOI & $n$ & Spearman $\rho$ & $p$ \\
\midrule
10.1016/j.seppur.2023.124580 & 13 & $-0.901$ & 0.000$^{*}$ \\
10.1021/jacsau.1c00414 & 9 & $-0.843$ & 0.004$^{*}$ \\
10.1016/j.micromeso.2015.01.044 & 13 & $-0.643$ & 0.018$^{*}$ \\
10.1016/j.jcou.2018.07.007 & 84 & $-0.565$ & 0.000$^{*}$ \\
10.1021/acs.chemmater.9b01474 & 11 & $-0.562$ & 0.072 \\
10.1021/acssuschemeng.0c03706 & 17 & $-0.547$ & 0.023$^{*}$ \\
10.1007/s10450-019-00171-w & 12 & $-0.523$ & 0.081 \\
10.1016/j.energy.2020.119093 & 47 & $-0.505$ & 0.000$^{*}$ \\
10.1021/acsapm.9b00788 & 8 & $-0.439$ & 0.276 \\
10.1021/acs.iecr.2c01484 & 24 & $-0.279$ & 0.186 \\
10.1021/acsami.8b12182 & 18 & $-0.216$ & 0.390 \\
10.1016/j.cej.2021.131875 & 22 & $-0.198$ & 0.377 \\
10.1021/ie300180u & 10 & $-0.182$ & 0.614 \\
10.1016/j.cej.2018.11.064 & 16 & $-0.109$ & 0.687 \\
10.1016/j.jece.2021.106709 & 29 & $-0.086$ & 0.656 \\
10.1016/j.seppur.2024.126735 & 90 & $-0.083$ & 0.438 \\
10.1016/j.seppur.2023.124346 & 17 & $+0.089$ & 0.734 \\
10.1021/acsami.5b07545 & 20 & $+0.128$ & 0.590 \\
10.1016/j.cej.2024.150025 & 36 & $+0.430$ & 0.009$^{*}$ \\
\bottomrule
\end{tabular}
\end{table}

\textbf{Data support at the 400 ppm operating point.} The $<1$ vol\% stratum of Table~\ref{tab:s_co2strata} spans concentrations up to $25\times$ the DAC operating point, so we audit the local support at 400 ppm directly (exp\_15\_ppm400\_support.json). The operating point coincides with the most densely sampled condition in the entire library: 169 of the 173 records in the $<1$ vol\% band, and 143 of the 146 records in the strict DAC window (20--35~°C, $\leq0.06$ vol\%), are measured at exactly 0.04 vol\% CO$_2$, spanning 21--22 publications with complete capacity and kinetics; library-wide, 0.04 vol\% measurements cover $-20$ to $75$~°C, and a single 0.02 vol\% record bounds the window from below. Predictions at 400 ppm are therefore interpolation at the modal concentration, not extrapolation, and the DAC-band errors reported above are dominated by exactly these records.

\textbf{Data support for the recommended design window.} The recommended candidate region ($45$--$55$ wt\%, $\hat{\phi}=0.4$--$0.5$) is a model-derived candidate region, and we audit its direct data support explicitly (exp\_17\_window\_stability.json). Support inside the narrow window is sparse by construction: 20 of the 618 library records (3.2\%), and 4 of the 146 strict-DAC-window records, fall inside it. The window is not a compositional artifact of the benchmark: the two amine-topology classes contribute exactly 10 in-window records each, and in both classes the in-window median capacity exceeds the out-of-window median (branched $2.67$ vs.\ $1.64$; linear $2.22$ vs.\ $1.67$ mmol\,g$^{-1}$; overall $2.41$ vs.\ $1.66$). In-window records concentrate in high-pore-volume hosts (MCF 6, MIL-101(Cr) 5), with a handful in BHMS/MMSN/THMS (2--4 records each); groups with fewer than 10 records overall or fewer than 5 in-window records are flagged \texttt{insufficient\_evidence} in the audit and are not cited for or against the window. SBA-15 contributes none of its 187 records, consistent with the window targeting high-pore-volume silicas rather than the literature's home turf. This sparsity is exactly why the main text frames the region as pending experimental validation.

\begin{table}[h]
\centering
\caption{Publication-grouped 10-fold cross-validation (GroupKFold by DOI) of the benchmark surrogate. Same columns as Table~\ref{tab:s_co2strata}; the gap between the two tables quantifies the interpolation--generalization gap.}\label{tab:s_groupedcv}
\begin{tabular}{lcccccc}
\toprule
Band & $n$ & \multicolumn{2}{c}{Capacity} & \multicolumn{2}{c}{Kinetics} \\
\cmidrule(lr){3-4}\cmidrule(lr){5-6}
 & & MAE & $R^2$ & MAE & $R^2$ \\
\midrule
DAC window ($<1$ vol\%) & 173 & 0.59 & 0.33 & 0.073 & $-13.2$ \\
Mid concentration (1--100 vol\%) & 92 & 0.70 & 0.50 & 0.136 & 0.15 \\
Pure CO$_2$ (100 vol\%) & 353 & 0.79 & 0.36 & 0.264 & $-0.04$ \\
\midrule
Overall & 618 & 0.72 & 0.45 & 0.191 & 0.13 \\
\bottomrule
\end{tabular}
\end{table}

\textbf{Oracle-constraint ablation (GP filter vs.\ analytic $c(x)$).} The constrained arm of the main campaign filters qLogEHVI proposals with a GP trained on noisy feasibility observations. Because the benchmark's feasibility ground truth is itself the closed form $c(x)=\tanh(1-\hat{\phi})-\tau$, we can measure how much filter fidelity matters: an oracle arm replaces the GP posterior-mean pass/fail with the noiseless analytic value, keeping every other component identical (same seeds, acquisition function, and budget; 100 seeds; Table~\ref{tab:s_oracle}). Final capacity is statistically indistinguishable ($3.029\pm0.821$ vs.\ $2.964\pm0.728$ mmol\,g$^{-1}$; paired Wilcoxon $p=0.60$), so no main-text conclusion hinges on filter fidelity. The filters do differ mechanically: per proposed candidate, the GP filter and the analytic constraint agree 59.1\% of the time, and the GP passes 88.3\% of proposals versus 52.2\% truly feasible---the GP smooths the sharp tanh transition zone and admits more boundary-region candidates. Those false passes are where infeasible front members come from: the oracle arm's front infeasible fraction drops to $7.9\pm17.3\%$ from $39.3\pm28.0\%$ (paired Wilcoxon $p=6.0\times10^{-13}$). Two consequences follow. First, the main-text $39.3\%$ figure is a conservative upper bound attributable to filter approximation error; the constraint itself is nearly clean. Second, the GP filter is the honest model of deployment, where $c(x)$ is not observable in closed form (pore volume is frequently unreported) and feasibility must be learned from noisy measurements---the oracle arm bounds what perfect constraint knowledge would add: deployability, not capacity. Hypervolume under this two-arm comparison ($1.244\pm0.584$ oracle vs.\ $1.158\pm0.436$ GP filter; shared reference point $[-0.391, -0.114]$ pooled over these two arms only) is not comparable to Table~\ref{tab:s_feashv}, which uses the four-method pooled reference point.

\begin{table}[h]
\centering
\caption{Oracle-constraint ablation (100 seeds, main-campaign budget). The GP-filter row is the constrained arm of the main campaign; the oracle row replaces the learned filter with the noiseless analytic constraint. Proposal pass rate is the fraction of qLogEHVI-proposed candidates admitted by each filter. $p$-values: paired Wilcoxon, oracle vs.\ GP filter.}\label{tab:s_oracle}
\begin{tabular}{lccc}
\toprule
Variant & Final capacity & Front infeasible & Proposal pass rate \\
\midrule
Constrained (GP filter) & $2.964\pm0.728$ & $39.3\pm28.0\%$ & 88.3\% \\
Oracle (analytic $c(x)$) & $3.029\pm0.821$ & $7.9\pm17.3\%$ & 52.2\% \\
\midrule
$p$ (oracle vs.\ GP filter) & $0.60$ (n.s.) & $6.0\times10^{-13}$ & -- \\
\bottomrule
\end{tabular}
\end{table}

\textbf{Classification quality of the feasibility GP.} The oracle ablation shows where the GP filter errs but not how well it classifies. Under the same publication-grouped 10-fold split, treating $c(x)\geq0$ as the true label and the GP posterior mean as the score (Table~\ref{tab:s_feascls}), the filter attains precision 0.933 and recall 0.935 across the full library (ROC-AUC 0.866; Brier score 0.109 against an 84.5\% base rate). Its full-library false-positive rate (36.5\%) confirms that over-admission concentrates outside the data-dense regions---the boundary smoothing diagnosed in the oracle ablation. Inside the DAC window the same filter is substantially sharper: precision 0.986, recall 0.933, false-positive rate 8.3\% (2 false passes in 173 records; ROC-AUC 0.975). The constraint is therefore most reliable exactly at the deployment-relevant conditions, and its residual DAC-window errors are conservative in direction: missed feasible candidates (10) outnumber false admissions (2).

\begin{table}[h]
\centering
\caption{Publication-grouped 10-fold CV classification quality of the feasibility GP at its operating point (posterior mean $\geq0$), against the analytic pore-filling boundary $c(x)\geq0$. FPR: false-positive rate (infeasible records passed). The DAC-window row is the deployment-relevant subset.}\label{tab:s_feascls}
\begin{tabular}{lcccccc}
\toprule
Region & $n$ & Precision & Recall & FPR & F$_1$ & ROC-AUC \\
\midrule
Full library & 618 & 0.933 & 0.935 & 0.365 & 0.934 & 0.866 \\
DAC window ($<1$ vol\%) & 173 & 0.986 & 0.933 & 0.083 & 0.959 & 0.975 \\
\bottomrule
\end{tabular}
\end{table}

\section{Feasible hypervolume: definitions and design choices}\label{sec:s5}

This section collects the precise construction of the deployability-aware metric used in the main text, and the rationale for each design choice. \textbf{Candidate set.} Standard hypervolume is computed over the objective vectors of \emph{all} candidates a method evaluated during a run (history points excluded); feasible hypervolume (feasible-HV) is computed over the subset satisfying $c(x)\geq0$:
\begin{equation}
\mathrm{HV}_{\mathrm{feas}}=\mathrm{HV}\bigl(\{f(x_i):c(x_i)\geq0\}\bigr),
\end{equation}
with $f$ the (capacity, kinetics) objective vector. The inflation ratio reported in the main text is $(\mathrm{HV}-\mathrm{HV}_{\mathrm{feas}})/\mathrm{HV}$ per run, averaged over seeds. \textbf{Shared reference point.} Both metrics use a single reference point $[-0.483, -0.196]$ computed once from the pooled objective vectors of all methods and seeds, so that HV differences between methods reflect the dominated region and not per-method reference-point choices; per-method reference points would confound the comparison we care about. \textbf{Full evaluated set, not front-only filtering.} Filtering only the final non-dominated front would credit a method for its feasible survivors while hiding the budget spent proposing infeasible candidates that were later discarded. Computing feasible-HV over the full evaluated set instead prices that budget: a method that spends half its evaluations outside the feasible region is penalized even if its surviving front looks clean. This is the property that makes the metric \emph{deployability-aware} rather than a post-hoc relabelling of the front. \textbf{Feasibility label and scope.} The feasible subset uses the analytic constraint value $c(x_i)$ of each \emph{evaluated} candidate (available in closed form for the benchmark); the learned GP filter only gates which proposals enter the evaluated set and does not enter the metric. Feasible-HV therefore prices infeasible candidates that consumed evaluation budget; it does not price proposals rejected before evaluation (which never consumed experimental budget), and it is not a complete deployability cost metric---it complements the evaluation-budget and front-composition readings reported alongside it. \textbf{Batch-size fairness.} Under the constraint, batches may contain fewer than 4 candidates because rejected proposals are not replaced (Methods). Both metrics are computed on evaluated candidates only, so this asymmetry enters through the search budget---the constrained optimizer simply has fewer evaluations in some iterations---and not through any normalization inside the metric; no per-iteration size correction is applied to either method. Quantitatively, the asymmetry is substantial: over 100 seeds the constrained arm spends $42.9\pm9.9$ true objective evaluations per run (median 43, range 23--64), versus exactly 80 (10 initial + $20\times4$) for every unconstrained method---the constrained optimizer is priced at roughly 54\% of the baselines' evaluation budget. We stress that this asymmetry is a property of constrained deployment rather than an accounting artifact: a rejected proposal is never synthesized or measured, so it consumes no experimental budget. The equal-evaluation-budget control below (d9) verifies that no feasible-HV conclusion depends on it. \textbf{Why a ratio, not a single metric.} Reporting standard HV and feasible-HV side by side separates two questions a single scalar would conflate: how much Pareto progress a method makes, and how much of that progress is realizable. We therefore treat the pair (plus the inflation ratio) as the reporting unit throughout.

\begin{figure}[t]
\centering
\includegraphics[width=0.75\textwidth]{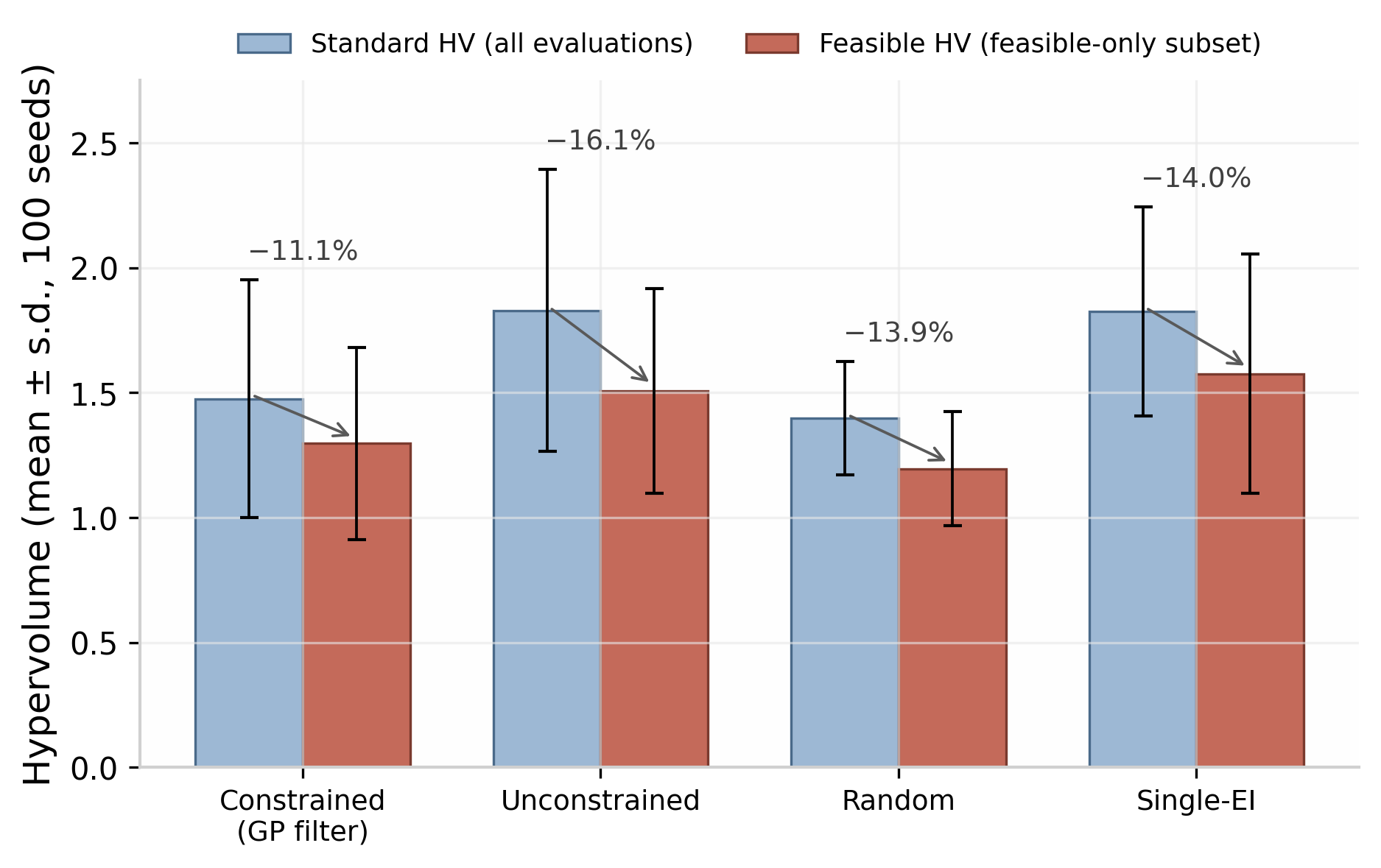}
\caption{\textbf{Standard versus feasible hypervolume for the four methods of the main campaign} (100 seeds; mean $\pm$ s.d.; shared reference point $[-0.483, -0.196]$). Arrows mark each method's inflation ratio, the fraction of its standard hypervolume carried by geometrically inconsistent candidates. Single-EI's high feasible-HV reflects its capacity-only objective, not deployability: 59.6\% of its Pareto-front candidates remain geometrically infeasible (main text). Numerical values: Table~\ref{tab:s_feashv}.}\label{fig:s_feashv}
\end{figure}

\textbf{100-seed values.} Table~\ref{tab:s_feashv} and Fig.~\ref{fig:s_feashv} report both metrics over the full 100-seed main campaign, recomputed from the per-seed evaluation histories with the shared reference point. The inflation gap between the constrained optimizer and unconstrained MOBO is significant (paired Wilcoxon on per-seed inflation, $p=2.1\times10^{-3}$); unconstrained MOBO's lead over the constrained optimizer narrows from 19\% under standard HV to 14\% under feasible-HV.

\textbf{Equal-evaluation-budget control (d9).} To verify that the budget asymmetry above does not drive these conclusions, we reran the constrained optimizer under an identical protocol but with the stopping rule changed to exactly 80 true objective evaluations (the unconstrained budget): proposals still pass through the same GP-mean filter and rejected candidates are still never evaluated; the arm simply iterates until the budget is spent (30 seeds, seeds 0--29, matching the robustness protocol of Section~\ref{sec:s6}; Table~\ref{tab:s_d9}). Two observations. First, every qualitative conclusion is unchanged at budget parity: unconstrained MOBO on the same 30 seeds still attains higher nominal capacity ($3.698\pm0.791$ vs.\ $3.091\pm0.804$ mmol\,g$^{-1}$, paired Wilcoxon $p=6.1\times10^{-4}$) at more than half again the front infeasible fraction ($67.8\%$ vs.\ $42.1\%$), and the equal-budget inflation ratio ($10.1\pm11.2\%$) matches the constrained arm's 100-seed value ($11.1\%$, $p=0.41$ between the two constrained variants) and remains significantly below unconstrained MOBO's ($18.5\pm13.9\%$ on these seeds, $p=0.003$). Second, the budget asymmetry was if anything \emph{conservative} for the constrained arm: with the full evaluation budget its capacity rises significantly over its own reject-without-replacement runs on the same seeds ($2.869\pm0.759$, paired Wilcoxon $p=7\times10^{-4}$)---the main-campaign comparison under-credits, rather than over-credits, the constraint.

\begin{table}[h]
\centering
\caption{Equal-evaluation-budget control (d9), 30 seeds (0--29), all arms evaluated on the same seeds. The d1 constrained arm spends $42.9\pm9.9$ true evaluations per run (reject without replacement); d9 spends exactly 80, matching the unconstrained budget. $p$-values: paired Wilcoxon against the d9 row.}\label{tab:s_d9}
\begin{tabular}{lccccc}
\toprule
Arm (30 seeds) & Evals & Final capacity & Front infeasible & Inflation \\
\midrule
Constrained, reject-no-replace (d1) & $42.9\pm9.9$ & $2.869\pm0.759$ & $43.2\%$ & $11.0\pm11.4\%$ \\
Constrained, equal budget (d9) & 80 & $3.091\pm0.804$ & $42.1\%$ & $10.1\pm11.2\%$ \\
Unconstrained (d1) & 80 & $3.698\pm0.791$ & $67.8\%$ & $18.5\pm13.9\%$ \\
\midrule
$p$ vs.\ d9 & -- & $7\times10^{-4}$ / $6.1\times10^{-4}$ & -- & $0.41$ / $0.003$ \\
\bottomrule
\end{tabular}
\end{table}

\begin{table}[h]
\centering
\caption{Standard vs.\ feasible hypervolume over 100 seeds (shared reference point $[-0.483, -0.196]$; inflation $=(\mathrm{HV}-\mathrm{HV}_{\mathrm{feas}})/\mathrm{HV}$ per run, mean $\pm$ s.d.\ across seeds).}\label{tab:s_feashv}
\begin{tabular}{lccc}
\toprule
Method & HV & Feasible-HV & Inflation \\
\midrule
Constrained MOBO & $1.476\pm0.475$ & $1.297\pm0.386$ & $11.1\pm12.3\%$ \\
Unconstrained MOBO & $1.829\pm0.564$ & $1.508\pm0.410$ & $16.1\pm13.3\%$ \\
Single-objective EI & $1.827\pm0.419$ & $1.576\pm0.480$ & $14.0\pm13.0\%$ \\
Random search & $1.397\pm0.227$ & $1.196\pm0.228$ & $13.9\pm11.1\%$ \\
\bottomrule
\end{tabular}
\end{table}

\section{Encoding, support-family, and constraint-form robustness}\label{sec:s6}

Three further checks probe whether the main-text conclusions depend on implementation choices. All use 30 seeds (seeds 0--29) at the main-campaign budget, and all keep the literature-calibrated ground truth fixed, so that each variant optimizes the same function as the main campaign; the ordinal-encoding, full-space reference on the same 30 seeds is included for like-for-like comparison (Table~\ref{tab:s_robust}). These variants were frozen at the pre-extension campaign size of 30 seeds; the main campaign was later extended to 100 seeds (Supplementary Section S4), and the ordinal reference here is the seed-matched subset of that extension, so no cross-seed-count comparison enters any conclusion.

\textbf{(a) One-hot encoding.} The main campaign ordinal-encodes categorical variables (10-dimensional optimizer input). Re-running the core comparison with one-hot encoding (66-dimensional optimizer-side representation), unconstrained MOBO retains higher nominal capacity ($3.731\pm1.003$ vs.\ $3.193\pm0.756$ mmol\,g$^{-1}$; paired Wilcoxon $p=0.004$) and a higher front infeasible fraction ($51.5\pm40.4\%$ vs.\ $32.5\pm30.9\%$); the constraint again shifts the front to lower loadings ($57.7$ vs.\ $63.4$ wt\%) and larger pores ($27.2$ vs.\ $22.4$ nm). Absolute infeasible fractions sit lower for \emph{both} optimizers than under ordinal encoding (the higher-dimensional surrogate extrapolates less aggressively in the high-loading region), but the gap between optimizers and every qualitative conclusion are encoding-invariant.

\textbf{(b) Support-family stratification (SBA-15-only).} Restricting candidate generation to SBA-15---the most densely sampled support family in the benchmark---the ordering persists: unconstrained $3.036\pm0.566$ vs.\ constrained $2.392\pm0.509$ mmol\,g$^{-1}$ ($p=1.1\times10^{-5}$), with front infeasible fractions of $83.3\pm21.8\%$ vs.\ $59.1\pm32.8\%$. Both fractions are higher than in the full-space campaign because fixing the support removes the optimizer's option of escaping to larger-pore hosts, sharpening contact with the boundary. The constrained front centres at 53.7 wt\% loading---inside the main-text deployable window---versus 67.5 wt\% unconstrained. The main conclusions are therefore not an artifact of support mixing.

\textbf{(c) Constraint form.} Replacing the smooth tanh feasibility target with binary sign labels ($\pm1$; the mean$\geq0$ filter is unchanged), the constrained optimizer attains a statistically indistinguishable capacity ($3.117\pm0.867$ vs.\ $2.869\pm0.759$ for the smooth form on the same seeds; $p=0.16$) with a comparable front infeasible fraction ($48.7\pm29.1\%$ vs.\ $43.2\pm26.7\%$; unconstrained: $67.8\pm27.6\%$). The smooth form is mildly cleaner---it retains gradient information across the transition zone, by design---but no conclusion depends on the choice.

\begin{table}[h]
\centering
\caption{Robustness variants (30 seeds each, seeds 0--29, main-campaign budget; ground truth fixed). The ordinal reference is the same-seed subset of the 100-seed main campaign. Wilcoxon $p$-values pair constrained vs.\ unconstrained within each variant (constraint-form row: binary vs.\ smooth constrained). The binary-feasibility variant has no unconstrained row: the unconstrained optimizer never queries the feasibility model, so its unconstrained behaviour is identical to the ordinal unconstrained reference row.}\label{tab:s_robust}
\footnotesize
\setlength{\tabcolsep}{3pt}
\begin{tabular}{llcccc}
\toprule
Variant & Group & Final capacity & Front infeasible & Loading (wt\%) & Pore (nm) \\
\midrule
Ordinal (reference) & constrained & $2.869\pm0.759$ & $43.2\pm26.7\%$ & 56.4 & 38.6 \\
                    & unconstrained & $3.698\pm0.791$ & $67.8\pm27.6\%$ & 66.1 & 28.8 \\
One-hot encoding    & constrained & $3.193\pm0.756$ & $32.5\pm30.9\%$ & 57.7 & 27.2 \\
                    & unconstrained & $3.731\pm1.003$ & $51.5\pm40.4\%$ & 63.4 & 22.4 \\
SBA-15-only         & constrained & $2.392\pm0.509$ & $59.1\pm32.8\%$ & 53.7 & 45.4 \\
                    & unconstrained & $3.036\pm0.566$ & $83.3\pm21.8\%$ & 67.5 & 32.8 \\
Binary feasibility  & constrained & $3.117\pm0.867$ & $48.7\pm29.1\%$ & 60.4 & 35.4 \\
\bottomrule
\end{tabular}
\end{table}


\begin{thebibliography}{99}

\bibitem{sanzperez2016}
Sanz-P\'erez, E. S., Murdock, C. R., Didas, S. A. \& Jones, C. W. Direct capture of CO$_2$ from ambient air. \emph{Chem. Rev.} \textbf{116}, 11840--11876 (2016).

\bibitem{jones2011}
Jones, C. W. CO$_2$ capture from dilute gases as a component of modern global carbon management. \emph{Annu. Rev. Chem. Biomol. Eng.} \textbf{2}, 31--52 (2011).

\bibitem{bui2018}
Bui, M. \emph{et al.} Carbon capture and storage (CCS): the way forward. \emph{Energy Environ. Sci.} \textbf{11}, 1062--1176 (2018).

\bibitem{keith2018}
Keith, D. W., Holmes, G., St.~Angelo, D. \& Heidel, K. A process for capturing CO$_2$ from the atmosphere. \emph{Joule} \textbf{2}, 1573--1594 (2018).

\bibitem{goeppert2011}
Goeppert, A. \emph{et al.} Carbon dioxide capture from the air using a polyamine based regenerable solid adsorbent. \emph{J. Am. Chem. Soc.} \textbf{133}, 20164--20167 (2011).

\bibitem{choi2011}
Choi, S., Drese, J. H., Eisenberger, P. M. \& Jones, C. W. Application of amine-tethered solid sorbents for direct CO$_2$ capture from the ambient air. \emph{Environ. Sci. Technol.} \textbf{45}, 2420--2427 (2011).

\bibitem{gebald2011}
Gebald, C., Wurzbacher, J. A., Tingaut, P., Zimmermann, T. \& Steinfeld, A. Amine-based nanofibrillated cellulose as adsorbent for CO$_2$ capture from air. \emph{Environ. Sci. Technol.} \textbf{45}, 9101--9108 (2011).

\bibitem{shi2020}
Shi, X. \emph{et al.} Sorbents for the direct capture of CO$_2$ from ambient air. \emph{Angew. Chem. Int. Ed.} \textbf{59}, 6984--7006 (2020).

\bibitem{didas2015}
Didas, S. A., Choi, S., Chaikittisilp, W. \& Jones, C. W. Amine--oxide hybrid materials for CO$_2$ capture from ambient air. \emph{Acc. Chem. Res.} \textbf{48}, 2680--2687 (2015).

\bibitem{bollini2011}
Bollini, P., Didas, S. A. \& Jones, C. W. Amine--oxide hybrid materials for acid gas separations. \emph{J. Mater. Chem.} \textbf{21}, 15100--15120 (2011).

\bibitem{sayari2010}
Sayari, A. \& Belmabkhout, Y. Stabilization of amine-containing CO$_2$ adsorbents: dramatic effect of water vapor. \emph{J. Am. Chem. Soc.} \textbf{132}, 6312--6314 (2010).

\bibitem{serna2010}
Serna-Guerrero, R., Belmabkhout, Y. \& Sayari, A. Further investigations of CO$_2$ capture using triamine-grafted pore-expanded mesoporous silica. \emph{Chem. Eng. J.} \textbf{158}, 513--519 (2010).

\bibitem{li2010}
Li, W. \emph{et al.} Steam-stripping for regeneration of supported amine-based CO$_2$ adsorbents. \emph{ChemSusChem} \textbf{3}, 899--903 (2010).

\bibitem{shahriari2016}
Shahriari, B., Swersky, K., Wang, Z., Adams, R. P. \& de~Freitas, N. Taking the human out of the loop: a review of Bayesian optimization. \emph{Proc. IEEE} \textbf{104}, 148--175 (2016).

\bibitem{frazier2018}
Frazier, P. I. A tutorial on Bayesian optimization. Preprint at \url{https://arxiv.org/abs/1807.02811} (2018).

\bibitem{daulton2020}
Daulton, S., Balandat, M. \& Bakshy, E. Differentiable expected hypervolume improvement for parallel multi-objective Bayesian optimization. In \emph{Advances in Neural Information Processing Systems} \textbf{33} (2020).

\bibitem{gelbart2014}
Gelbart, M. A., Snoek, J. \& Adams, R. P. Bayesian optimization with unknown constraints. In \emph{Proc. 30th Conference on Uncertainty in Artificial Intelligence} (2014).

\bibitem{gramacy2011}
Gramacy, R. B. \& Lee, H. K. H. Optimization under unknown constraints. \emph{Bayesian Statistics} \textbf{9} (2011).

\bibitem{gardner2014}
Gardner, J. R., Kusner, M. J., Xu, Z., Weinberger, K. Q. \& Cunningham, J. P. Bayesian optimization with inequality constraints. In \emph{Proc. 31st International Conference on Machine Learning} (2014).

\bibitem{letham2019}
Letham, B., Karrer, B., Ottoni, G. \& Bakshy, E. Constrained Bayesian optimization with noisy experiments. \emph{Bayesian Anal.} \textbf{14}, 495--519 (2019).

\bibitem{eriksson2021}
Eriksson, D. \& Poloczek, M. Scalable constrained Bayesian optimization. In \emph{Proc. 24th International Conference on Artificial Intelligence and Statistics} (2021).

\bibitem{balandat2020}
Balandat, M. \emph{et al.} BoTorch: a framework for efficient Monte-Carlo Bayesian optimization. In \emph{Advances in Neural Information Processing Systems} \textbf{33} (2020).

\bibitem{ament2023}
Ament, S., Daulton, S., Eriksson, D., Balandat, M. \& Bakshy, E. Unexpected improvements to expected improvement for Bayesian optimization. In \emph{Advances in Neural Information Processing Systems} \textbf{36} (2023).

\bibitem{wang2025scidata}
Wang, E., Luo, L., Wang, J., Dai, J., Li, S., Chen, L. \& Li, J. A dataset for investigations of amine-impregnated solid adsorbent for direct air capture. \emph{Sci. Data} \textbf{12}, 724 (2025).

\bibitem{berger2011}
Berger, A. H. \& Bhown, A. S. Comparing physisorption and chemisorption solid sorbents for use separating CO$_2$ from flue gas using temperature swing adsorption. \emph{Energy Procedia} \textbf{4}, 562--567 (2011).

\bibitem{deb2002}
Deb, K., Pratap, A., Agarwal, S. \& Meyarivan, T. A fast and elitist multiobjective genetic algorithm: NSGA-II. \emph{IEEE Trans. Evol. Comput.} \textbf{6}, 182--197 (2002).

\bibitem{blank2020}
Blank, J. \& Deb, K. pymoo: multi-objective optimization in Python. \emph{IEEE Access} \textbf{8}, 89497--89509 (2020).

\bibitem{mcdonald2015}
McDonald, T. M. \emph{et al.} Cooperative insertion of CO$_2$ in diamine-appended metal--organic frameworks. \emph{Nature} \textbf{519}, 303--308 (2015).

\bibitem{gardner2018}
Gardner, J. R., Pleiss, G., Bindel, D., Weinberger, K. Q. \& Wilson, A. G. GPyTorch: blackbox matrix--matrix Gaussian process inference with GPU acceleration. In \emph{Advances in Neural Information Processing Systems} \textbf{31} (2018).

\bibitem{zitzler1999}
Zitzler, E. \& Thiele, L. Multiobjective evolutionary algorithms: a comparative case study and the strength Pareto approach. \emph{IEEE Trans. Evol. Comput.} \textbf{3}, 257--271 (1999).

\bibitem{while2006}
While, L., Hingston, P., Barone, L. \& Huband, S. A faster algorithm for calculating hypervolume. \emph{IEEE Trans. Evol. Comput.} \textbf{10}, 29--38 (2006).

\bibitem{seabold2010}
Seabold, S. \& Perktold, J. Statsmodels: econometric and statistical modeling with Python. In \emph{Proc. 9th Python in Science Conference} (2010).

\bibitem{pedregosa2011}
Pedregosa, F. \emph{et al.} Scikit-learn: machine learning in Python. \emph{J. Mach. Learn. Res.} \textbf{12}, 2825--2830 (2011).

\end{thebibliography}
\end{document}